%% file: main.tex
\pdfoutput=1
\documentclass[10pt,twocolumn,letterpaper]{article}

\usepackage[pagenumbers]{cvpr} 

\input{preamble}

\definecolor{cvprblue}{rgb}{0.21,0.49,0.74}
\usepackage[pagebackref,breaklinks,colorlinks,allcolors=cvprblue]{hyperref}
\usepackage{xcolor}
\usepackage{CJK}
\usepackage{amssymb}
\usepackage{wasysym}
\usepackage{amsmath}        
\usepackage{amsfonts}       
\usepackage{bm}             
\usepackage{enumitem}       
\usepackage{booktabs}       
\usepackage{colortbl}       
\usepackage{multirow}
\usepackage{makecell}
\usepackage{setspace}
\usepackage{graphicx} 
\usepackage{arydshln}
\usepackage[ruled,vlined,linesnumbered]{algorithm2e}
\usepackage{pifont}
\usepackage{tcolorbox}
\usepackage{float}
\tcbuselibrary{skins, breakable}

\def\paperID{1872} 
\def\confName{CVPR}
\def\confYear{2026}

\title{VCU-Bridge: Hierarchical Visual Connotation Understanding via Semantic Bridging
}
\newcommand{\topic}{VCU-Bridge}
\newcommand{\lperc}{$L_{perc}$}
\newcommand{\lbridge}{$L_{bridge}$}
\newcommand{\lconn}{$L_{conn}$}
\newcommand{\aperc}{$Acc_{perc}$}
\newcommand{\abridge}{$Acc_{bridge}$}
\newcommand{\aconn}{$Acc_{conn}$}
\newcommand{\afull}{$Acc_{full}$}
\newcommand{\bench}{HVCU-Bench}
\newcommand{\lora}{Qwen3-VL-4B-Bridge}
\definecolor{mypink}{RGB}{255,235,240}
\newcommand{\myspace}{0mm}
\author{
    Ming Zhong$^{1, *, \spadesuit}$ \quad
    Yuanlei Wang$^{3, *}$ \quad
    Liuzhou Zhang$^{2, *}$ \quad
    Ruichuan An$^{2, \spadesuit}$ \quad
    Renrui Zhang$^{4}$ \\ %
    Hao Liang$^{2}$ \quad
    Ming Lu$^{2}$ \quad
    Ying Shen$^{3, \dagger}$ \quad
    Wentao Zhang$^{2, \dagger}$
    \vspace{5pt} \\
    $^{1}$Zhejiang University \quad
    $^{2}$Peking University \quad
    $^{3}$Sun Yat-sen University \quad
    $^{4}$CUHK
    \vspace{5pt} \\ 
    {\tt\small Project page: \href{https://vcu-bridge.github.io}{https://vcu-bridge.github.io}}
}

\begin{document}
\maketitle
\let\thefootnote\relax\footnotetext{$^*$ Equal Contribution, Email: chime@zju.edu.cn}
\let\thefootnote\relax\footnotetext{$^\spadesuit$ Project Leader}
\let\thefootnote\relax\footnotetext{$^\dagger$ Corresponding Authors}

\input{sec/0_abstract}    
\input{image-and-table/case4motivation}
\input{image-and-table/motivation}
\input{image-and-table/benchmark_table}
\input{sec/1_intro}
\input{sec/2_related-work}

\input{sec/3_method}
\input{sec/4_experiments}
\input{sec/5_Conclusion}
{
    \small
    \bibliographystyle{ieeenat_fullname}
    \bibliography{main}
}
\input{sec/X_suppl}


\end{document}

%% file: preamble.tex
\renewcommand{\paragraph}[1]{\vspace{.6em}\noindent\textbf{#1}}



%% file: sec/0_abstract.tex
\begin{abstract}
While Multimodal Large Language Models (MLLMs) excel on benchmarks, their processing paradigm differs from the human ability to integrate visual information. Unlike humans who naturally bridge details and high-level concepts, models tend to treat these elements in isolation. Prevailing evaluation protocols often decouple low-level perception from high-level reasoning, overlooking their semantic and causal dependencies, which yields non-diagnostic results and obscures performance bottlenecks. We present \textbf{\textit{\topic{}}}, a framework that operationalizes a human-like hierarchy of visual connotation understanding: multi-level reasoning that advances from foundational perception through semantic bridging to abstract connotation, with an explicit evidence-to-inference trace from concrete cues to abstract conclusions. Building on this framework, we construct \textbf{\textit{\bench{}}}, a benchmark for hierarchical visual connotation understanding with explicit, level-wise diagnostics. Comprehensive experiments demonstrate a consistent decline in performance as reasoning progresses to higher levels. We further develop a data generation pipeline for instruction tuning guided by Monte Carlo Tree Search (MCTS) and show that strengthening low-level capabilities yields measurable gains at higher levels. Interestingly, it not only improves on \bench{} but also brings benefits on general benchmarks (average +2.53\%), especially with substantial gains on MMStar (+7.26\%), demonstrating the significance of the hierarchical thinking pattern and its effectiveness in enhancing MLLM capabilities.
\end{abstract}
\vspace{-3mm}

%% file: image-and-table/case4motivation.tex
\begin{figure}[t!]
    \centering
    \includegraphics[width=0.48\textwidth]{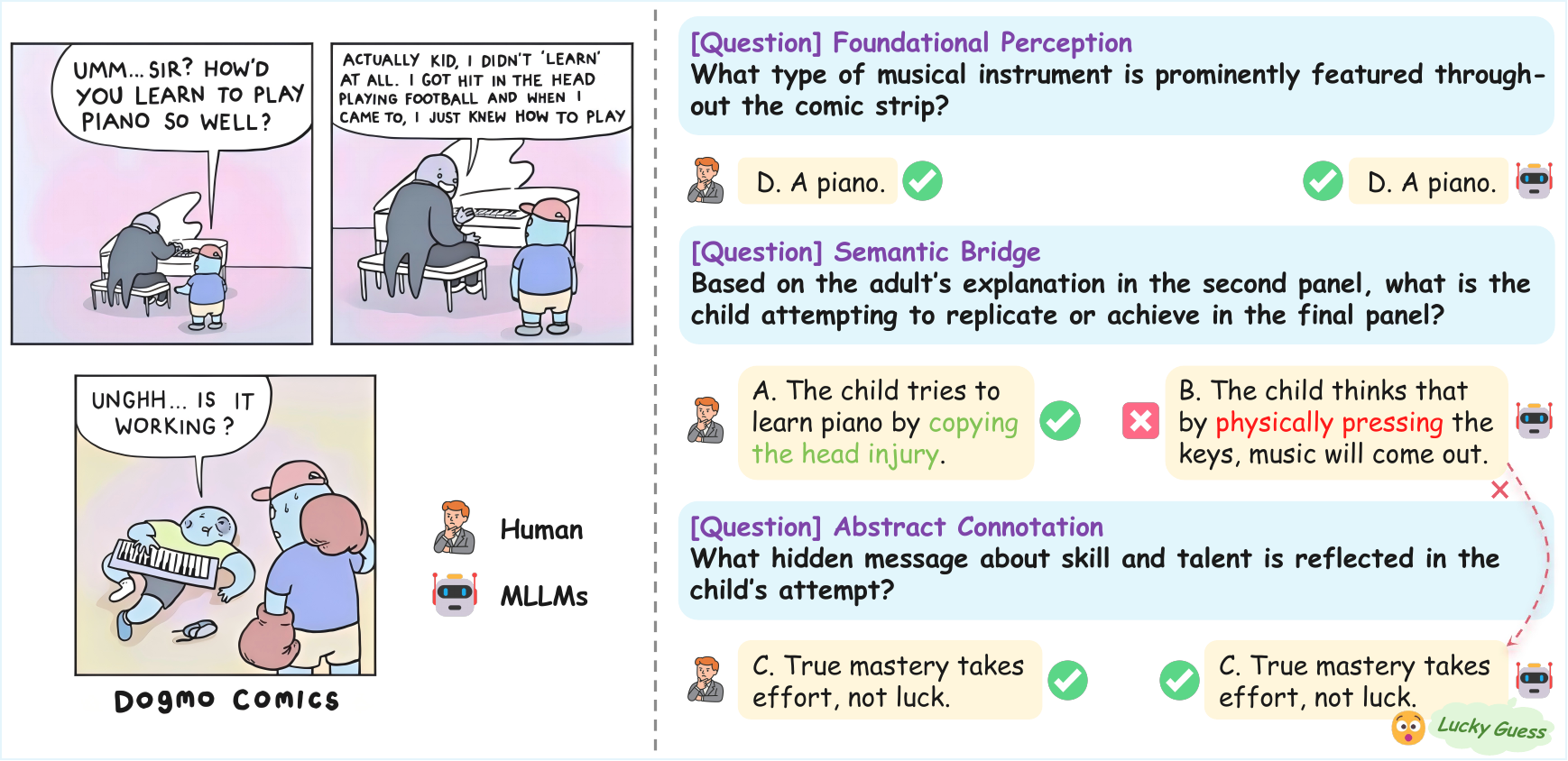}
    \caption{\textbf{Showcase of different pattern between human and model.} Models can appear capable by correctly answering both concrete and abstract questions while fundamentally failing at the reasoning that bridges them. Current evaluation may miss critical reasoning failures when models produce correct answers at both concrete and abstract levels.}
    \label{fig:motivation}
    \vspace{\myspace}
\end{figure}

%% file: image-and-table/motivation.tex
\begin{figure*}[!t]
    \centering
    \includegraphics[width=0.97\textwidth]{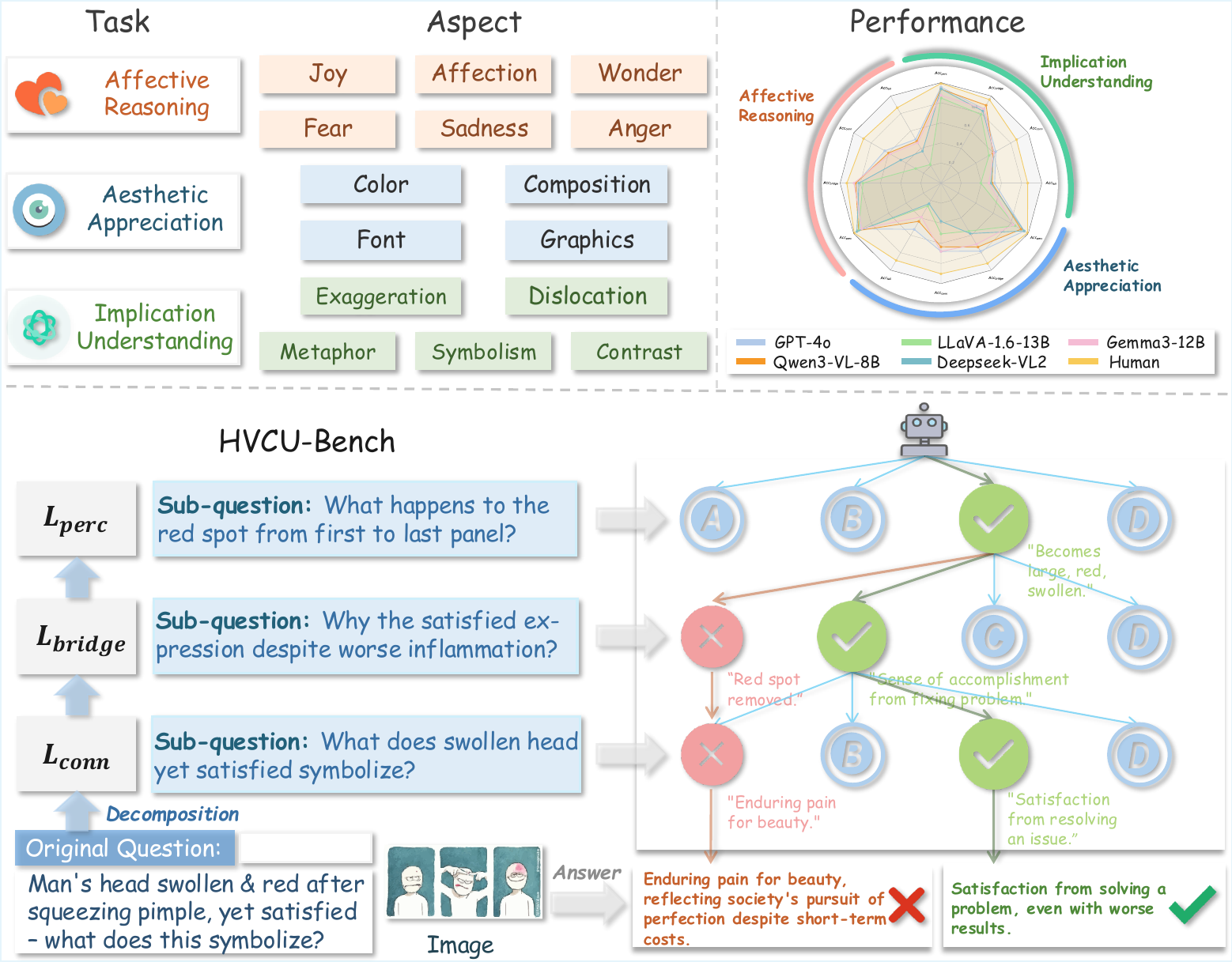}
    \caption{\textbf{Overview of \bench{}.} 
    We evaluate MLLMs across 3 task families spanning 15 diverse aspects (\textbf{top left}). Our benchmark employs hierarchical decomposition: each question is systematically broken down into sub-questions across three levels (\lperc{}, \lbridge{}, \lconn{}), with validation ensuring logical coherence. During evaluation, models progress from low to high levels, constructing inter-level reasoning chains that emulate human visual comprehension (\textbf{bottom}). While GPT-4o achieves top performance among MLLMs, it falls substantially short of human capability, exposing a significant gap (\textbf{top right}).}
    \label{fig:bench}
    \vspace{\myspace}
\end{figure*}

%% file: image-and-table/benchmark_table.tex
{\renewcommand{\arraystretch}{0.89}
\begin{table*}[t!]
\centering
\caption{\textbf{Comparison of various recent multimodal benchmarks about visual understanding.}}
\resizebox{\textwidth}{!}{
\begin{tabular}{lcccccccc}
\toprule
\multirow{3}{*}{\textbf{Benchmark}} &
\multicolumn{4}{c}{\textbf{\topic{}}} &
\multicolumn{2}{c}{\textbf{Metrics}} &
\multirow{3}{*}{\textbf{Human Evaluation}} &
\multirow{3}{*}{\textbf{Open-Source}} \\
\cmidrule(lr){2-5} 
\cmidrule(lr){6-7}
& \textbf{\makecell{Foundational \\ Perception}} & \textbf{\makecell{Semantic \\ Bridge}} & \textbf{\makecell{Abstract \\ Connotation}} & 
\textbf{\makecell{Inter-Level \\ Association}} & 
\textbf{Accuracy} & \textbf{\makecell{Full-Chain \\ Accuracy}} & 
& \\
\midrule
\multicolumn{9}{l}{\textbf{Low-level}} \\
\midrule
MME~\cite{fu2024mme}  & \checked & $\times$ & $\times$ & $\times$ & \checked & $\times$ & $\times$ & \checked \\
MMBench~\cite{liu2023mmbench}  & \checked & $\times$ & $\times$ & $\times$ & \checked & $\times$ & \checked & \checked \\
MMStar~\cite{mmstar}  & \checked & $\times$ & $\times$ & $\times$ & \checked & $\times$ & $\times$ & \checked \\
HallusionBench~\cite{guan2024hallusionbench}  & \checked & $\times$ & $\times$ & $\times$ & \checked & $\times$ & $\times$ & \checked \\
HCA~\cite{tan2025visionllmsbadhierarchical} & \checked & $\times$ & $\times$ & $\times$ & \checked & \checked & $\times$ & \checked \\
\midrule
\multicolumn{9}{l}{\textbf{High-level}} \\
\midrule
MMMU~\cite{mmmu}  & $\times$ & $\times$ & \checked & $\times$ & \checked & $\times$ & \checked & \checked \\
MME-Reasoning~\cite{he2025mmereasoning} & $\times$ & $\times$ & \checked & $\times$ & \checked & $\times$ & \checked & \checked \\
II-Bench~\cite{ii-bench}               & $\times$ & $\times$ & \checked & $\times$ & \checked & $\times$ & \checked & \checked \\
EEmo-Bench~\cite{eemo-bench}  & $\times$ & $\times$ & \checked & $\times$ & \checked & $\times$ & $\times$ & \checked \\
\midrule
\multicolumn{9}{l}{\textbf{Multi-level}} \\
\midrule
Lens~\cite{yao2025lens}  & \checked & $\times$ & \checked & $\times$ & \checked & $\times$ & $\times$ & \checked \\
MVP-Bench~\cite{mvp-bench}  & \checked & $\times$ & \checked & $\times$ & \checked & $\times$ & \checked & $\times$ \\
HiBench~\cite{hibench}             & \checked & $\times$ & \checked & $\times$ & \checked & $\times$ & $\times$ & $\times$ \\
InsightVision~\cite{insightvision}       & \checked & $\times$ & \checked & $\times$ & \checked & $\times$ & $\times$ & $\times$ \\
\midrule
\textbf{\bench{} (Ours)}   & \checked & \checked & \checked & \checked & \checked & \checked & \checked & \checked \\
\bottomrule
\end{tabular}
}
\label{tab:benchmark_comparison}
\vspace{\myspace}
\end{table*}
}

%% file: sec/1_intro.tex
\section{Introduction}
\label{sec:intro}

While Multimodal Large Language Models (MLLMs) excel on benchmarks, their processing paradigm differs from the human ability to integrate visual information.  As illustrated in Figure~\ref{fig:motivation}, unlike humans who naturally bridge details and high-level concepts, models tend to treat these elements in isolation. This phenomenon misses a vital aspect of human vision: our capacity to form abstract, connotative meanings based on concrete perceptual evidence~\cite{serafini2010reading}. The inferential process that connects seeing to interpreting remains largely unevaluated. 

Existing evaluation paradigms fall short in capturing this critical inferential process. Benchmarks focused on low-level perception~\cite{ge2025mllm-bench, qiang2025ver, zhu2025lime} test the ability to identify what is in an image, but do not evaluate the ability to infer what is implied by it. Conversely, benchmarks for high-level cognition~\cite{tang2025big, wang2024mementos, he2025mmereasoning, tang2025mtvqabenchmarkingmultilingualtextcentric} often assess the final abstract conclusion but do not require the model to ground its reasoning in specific visual details, leaving the inferential process a black box. What remains missing is an explicit focus on the bridging inference: the associative reasoning that connects concrete perceptual evidence to abstract connotative meaning (e.g., dark clouds and a hunched posture create a scene of a person alone in bad weather, which connotes a sense of gloom)~\cite{serafini2010reading,wildgen2024cognitive}, and addressing this critical gap is the primary motivation of our work.

To address this, we introduce and formalize the concept of \textbf{\textit{\topic{}}}, which refers to hierarchical understanding of visual connotation. We propose that effective visual comprehension operates as a hierarchical, multi-level reasoning process. It begins at the \textbf{Foundational Perceptual Level (\lperc{})}, which involves identifying low-level, objective visual primitives like objects and their attributes. It culminates in the \textbf{Abstract Connotative Level (\lconn{})}, where high-level, subjective interpretations such as aesthetics, emotion, or symbolic meaning are inferred. Critically, these two levels are connected by an intermediate \textbf{Semantic Bridge Level (\lbridge{})}, which provides factual, descriptive statements that explain how perceptual evidence gives rise to connotative meaning. It is this unmodeled ``Bridge'' level, which explicitly links the concrete and the abstract, that remains a key unsolved challenge for MLLMs. \topic{} is the first to focus on the connotative hierarchy, demanding that the model articulate the associative link from \lperc{}, through \lbridge{}, to \lconn{}. The core of our contribution is thus the modeling of this inter-level association, which serves as the foundation for evaluation.

To systematically explore and evaluate the \topic{} capabilities of MLLMs, we construct \textbf{\textit{\bench{}}}, a novel benchmark specifically designed to measure this hierarchical visual reasoning ability. \bench{} explicitly models the hierarchical process of visual understanding, requiring models to reason step-by-step from low-level perception to high-level abstraction, which forms tightly coupled inter-level reasoning chains. Through hierarchical evaluation, \bench{} enables tracing failures at higher levels back to potential causes in lower levels, thereby facilitating a more fine-grained analysis of model capability bottlenecks. Figure~\ref{fig:bench} provides an overview of \bench{}.

Subsequently, to enhance the hierarchical reasoning ability of models, we propose a targeted low-to-high-level data generation approach based on Monte Carlo Tree Search (MCTS)~\cite{li2025enhancing,  yang2025reranking, zhou2024language, luo2024llm, an2025conceptastree}. Specifically, we design a data generation pipeline that starts from low-level perceptual details, progressively builds up to high-level abstract questions, and validates each level to ensure the logical coherence and quality of the hierarchical reasoning chain. The MCTS strategy enables efficient exploration and generation of high-quality training data within the hierarchical reasoning space for instruction tuning.

In summary, our contributions are as follows:
\begin{itemize}
\item We novelly introduce and formalize \textbf{\textit{\topic{}}}, modeling visual connotation understanding as a three-level progressive process (\lperc{} / \lbridge{} / \lconn{}), and highlighting the importance of modeling inter-level associations.
\item We construct \textbf{\textit{\bench{}}}, the first benchmark to explicitly evaluate \topic{} capabilities. \bench{} enables hierarchical evaluation and diagnosis of deep causes for high-level reasoning failures, providing fine-grained analysis of model capability bottlenecks.
\item We propose a novel MCTS-based data generation pipeline that explores the hierarchical reasoning space, progressively constructing chains from low-level perceptual details to high-level abstract concepts with level-wise validation to ensure logical coherence and quality.
\item Extensive experiments on \bench{} reveal universal performance degradation from perception to connotation and substantial gains with lower-level context. Models instruction-tuned with data from our pipeline achieve significant improvements on \bench{} (+6.17\%) while maintaining strong performance on diverse general benchmarks (average +2.53\%).
\end{itemize}

%% file: sec/2_related-work.tex
\section{Related Work}
\label{sec:formatting}

\paragraph{Visual Understanding Evaluation.}
The evaluation of Multimodal Large Language Models (MLLMs) has advanced rapidly~\cite{an2024mc, lin2025perceiveanythingrecognizeexplain, lin2024draw} , but a significant gap persists between low-level perception and high-level connotation. Many existing benchmarks tend to isolate these capabilities, failing to address the inferential processes that link perception with connotation.
\textbf{Low-level Perception.} Benchmarks such as VQA~\cite{vqa} and MMBench~\cite{liu2023mmbench} test object recognition, spatial understanding, and factual grounding but do not evaluate the inferential leap from \textit{what is in} an image to \textit{what is implied} by it. Although HCA~\cite{tan2025visionllmsbadhierarchical} measures full-chain accuracy across hierarchical categories, it focuses on taxonomic classification relying on pretrained knowledge rather than genuine hierarchical visual understanding.
\textbf{High-level Connotation.} Benchmarks such as II-Bench~\cite{ii-bench} and EEmo-Bench~\cite{eemo-bench} assess logical inference and metaphorical interpretation from \textit{what is implied} but do not require grounding in \textit{what is in} the image.
\textbf{Multi-level.} Benchmarks such as Lens~\cite{yao2025lens} and MVP-Bench~\cite{mvp-bench} cover multiple levels but without inter-level associations, leaving the reasoning pathway from \textit{what is in} to \textit{what is implied} implicit. A comprehensive comparison between our benchmark and prior works is presented in Table \ref{tab:benchmark_comparison}.
Additional discussion on data generation approaches for instruction tuning is provided in Appendix~\ref{sec:more_related_works}.

%% file: sec/3_method.tex
\section{Methods}

\subsection{Problem Formulation}
We formalize \textbf{\textit{\topic{}}} as the structured inference of abstract scene interpretations grounded in perceptual evidence, with each intermediate semantic justification made explicit and auditable. Formally, \topic{} is represented as a structured triple of statements $\mathcal{H} = (l_1, l_2, l_3)$, ordered by their level of abstraction $l_1 \rightarrow l_2 \rightarrow l_3$. A valid hierarchy must satisfy a pairwise support constraint $S(l_k, l_{k+1})$ for $k\in\{1,2\}$, where $S$ indicates that the higher-level statement $l_{k+1}$ is sufficiently justified by the lower-level one $l_k$. The three levels are defined as follows:
\begin{itemize}
    \item \textbf{Foundational Perceptual Level (\lperc{} = $l_1$):} Objective, low-level visual facts that are directly observable in the image.
    \item \textbf{Semantic Bridge Level (\lbridge{} = $l_2$):} Explanatory statements that causally link perceptual evidence to higher-level meaning.
    \item \textbf{Abstract Connotative Level (\lconn{} = $l_3$):} Subjective, high-level interpretations of the scene.
\end{itemize}

This formulation motivates two subsequent components: (1) \textbf{\textit{\bench{}}}, which operationalizes the three-level structure and evaluates both per-level and full-chain performance; and (2) a targeted data generation pipeline that supplies high-quality hierarchical supervision to systematically and significantly improve MLLMs on \topic{}.

\subsection{\bench{}: A Benchmark for \topic{}}
\label{subsec:bench}
To systematically evaluate \topic{}, we introduce \bench{}, a benchmark constructed through a model-driven generation-validation framework powered by Gemini-2.5-Pro. The core principle is a sequential generation process from abstract to concrete, where each level conditions on the previous one, with immediate validation applied after generating each subsequent level to ensure logical coherence and hierarchical dependencies. This interleaved generation-validation approach produces question-answer pairs that are both visually grounded and exhibit strong hierarchical relationships. 
Detailed statistics, quality assurance procedures, and representative samples are provided in Appendix~\ref{sec:bench_details}. 

\paragraph{Task Design.}
\bench{} comprises three task families covering fifteen fine-grained aspects:
\textit{Affective Reasoning} (\textit{joy}, \textit{affection}, \textit{wonder}, \textit{anger}, \textit{fear}, \textit{sadness});
\textit{Aesthetic Appreciation} (\textit{color}, \textit{composition}, \textit{font}, \textit{graphics});
\textit{Implication Understanding} (\textit{metaphor}, \textit{symbolism}, \textit{contrast}, \textit{exaggeration}, \textit{dislocation}).
All items follow a hierarchical multiple-choice QA format with four options per question.

For Affective and Implication tasks, we leverage images and high-level concepts from existing datasets II-Bench~\cite{ii-bench} and EEmo-Bench~\cite{eemo-bench} to anchor the \lconn{} level, then systematically construct answer options and the lower levels (\lbridge{}, \lperc{}) through our generation framework.
For Aesthetic tasks, we source images from LayerD~\cite{suzuki2025layerd} and apply controlled perturbations for each aspect to isolate the targeted aesthetic factor while keeping the hierarchical reasoning structure consistent. For instance, color variants are created by adjusting hue, saturation, and brightness; composition variants by modifying cropping, reframing, and scale; font variants generated by changing size, weight, and spacing; diverse graphics variants generated by altering stroke width, texture, and density.

\paragraph{Sequential Generation with Interleaved Validation.}
Each benchmark sample is generated through a three-stage process, where each level $l_i$ is instantiated as a question–answer pair $(q_i, a_i)$. Given an input image $I$, the generation proceeds from abstract to concrete:

\begin{itemize}
    \item \textbf{Stage 1 - Connotative Synthesis (\lconn):} An MLLM performs holistic image analysis to generate the highest-level question–answer pair $(q_3, a_3)$, capturing abstract interpretations such as aesthetic quality or emotional tone.
    
    \item \textbf{Stage 2 - Semantic Bridge Formulation (\lbridge):} Conditioned on $(q_3, a_3)$, the model generates an intermediate-level pair $(q_2, a_2)$ that provides semantic justification for the connotative interpretation. Validation then checks whether $(q_2, a_2)$ offers sufficient logical support for $(q_3, a_3)$. If validation fails, a refinement loop is triggered to regenerate $(q_2, a_2)$.
    
    \item \textbf{Stage 3 - Perceptual Grounding (\lperc):} Conditioned on both $(q_3, a_3)$ and $(q_2, a_2)$, the model generates the most concrete pair $(q_1, a_1)$, grounding the reasoning chain in observable visual evidence. Similarly, validation ensures $(q_1, a_1)$ provides foundational support for $(q_2, a_2)$, triggering a refinement loop if validation fails.
\end{itemize}

\paragraph{Validation Mechanism and Refinement.}
The validation enforces two critical criteria: (1) \textit{logical dependency}—the lower-level answer must be directly useful for reasoning about the higher-level question, and (2) \textit{difficulty progression}—the lower level must be significantly more objective, concrete, and observable compared to the higher level's interpretative nature.

When validation fails, a refinement loop is triggered. The failed generation is discarded, and the stage re-executes with two adaptive strategies: (1) incorporating the specific validation failure reason as explicit guidance to more effectively avoid similar logical flaws, and (2) progressively increasing the sampling temperature to further enhance output diversity and escape potential local optima.
Prompts for generation and validation are provided in Appendix~\ref{sec:prompts}.

\paragraph{Evaluation Metrics.}
We evaluate model performance on \bench{} using metrics that capture both level-specific correctness and full-chain consistency. For each level $i$, let $a_i$ denote the ground-truth answer and $\hat{a}_i$ the model's prediction.
\begin{itemize}
    \item \textbf{Per-Level Accuracy ($Acc_i$):} The proportion of correct predictions at each individual level, formally $Acc_i = \mathbb{P}[\hat{a}_i = a_i]$ for $i \in \{1, 2, 3\}$, providing insight into capability at different abstraction levels.
    \item \textbf{Full-Chain Accuracy (\afull):} The most stringent metric, defined as \afull{} $= \mathbb{P}[\bigwedge_{i=1}^3 (\hat{a}_i = a_i)]$, requires simultaneous correctness across all three levels. This evaluates the model's ability to perform hierarchical reasoning from perception to connotation.
\end{itemize}
To provide a single aggregate performance indicator across all three tasks, we also report an Overall Score (\textit{Score}), computed as the mean of the \afull{} scores across all these tasks, enabling a fair, robust, and direct model comparison on overall \topic{} capability.

\subsection{Hierarchical Data Generation for Instruction Tuning}
\label{subsec:data_generation}

\input{image-and-table/overview}

To enhance the hierarchical reasoning capabilities of MLLMs, we propose a novel data generation pipeline for instruction tuning. In contrast to the benchmark's top-down sequential generation of individual samples, our training data is constructed through a large-scale bottom-up search. Rather than refining a single reasoning chain per image, we employ MCTS to explore a tree of candidate paths, ultimately selecting diverse high-quality chains for training. Figure~\ref{fig:pipeline} illustrates the complete pipeline.

\paragraph{MCTS-driven Reasoning Path Discovery.}
The core of our generation pipeline is an MCTS procedure that iteratively constructs a reasoning tree. Starting from a virtual root, the algorithm progressively expands question-answer pairs across the configured hierarchical levels, with each node representing a (q, a) pair at a specific reasoning level. The search balances exploration and exploitation via the Upper Confidence Bound (UCB) strategy, proceeding in three iterative phases:
\begin{itemize}
    \item \textbf{Selection:} The algorithm selects nodes to expand based on UCB: $UCB_j = \bar{X}_j + c \sqrt{\frac{\ln N}{n_j}}$, where $\bar{X}_j$ denotes node $j$'s average reward, $n_j$ its visit count, $N$ the parent's visits, and $c$ is the exploration constant controlling the exploration-exploitation trade-off. This effectively balances exploitation (high-reward nodes) and exploration (less-visited nodes), while prioritizing promising yet underexplored reasoning paths.
    \item \textbf{Expansion \& Evaluation:} For each selected node, the pipeline generates a batch of candidate children through parallel MLLM calls, with batch size constrained by a configurable capacity limit per expansion. Each candidate undergoes immediate quality assessment to obtain a numerical score (ranging from 0 to 1), with rejections occurring when candidates exhibit poor logical coherence with the parent, excessive semantic similarity to existing siblings, exceed per-level capacity limits, or fall below a quality threshold (e.g., insufficient image alignment or inappropriate difficulty level). Only candidates successfully passing these checks are admitted to the tree with their evaluation scores, ensuring a more robust balance between exploration breadth and reasoning validity.
    \item \textbf{Backpropagation:} Evaluation scores propagate back up the tree, updating statistics for all ancestors. Specifically, for each ancestor node $j$ on the path, we update $n_j \leftarrow n_j + 1$ and $\bar{X}_j \leftarrow \frac{(n_j - 1) \cdot \bar{X}_j + r}{n_j}$, where $r$ is the newly observed reward. This credit assignment ensures high-quality branches accumulate higher rewards, receiving more exploration in iterations.
\end{itemize}
The pipeline maintains a hierarchical tree structure with configurable depth and per-level node capacity limits. Through iterative MCTS cycles, combined with quality filtering and diversity checks, the search efficiently explores the reasoning space at scale, constructing a diverse tree of promising, high-quality candidate paths with significantly improved exploration-exploitation trade-offs.
Detailed prompts for node generation and evaluation are provided in Appendix~\ref{sec:prompts}.

\input{image-and-table/main_results}
\paragraph{Top-K Reasoning Path Selection.}
After MCTS convergence, the pipeline extracts the top-K highest-rated complete reasoning paths from the search tree, where each path traverses the configured hierarchical levels from low-level perceptual grounding to high-level abstract interpretation. Path ranking is based on the average evaluation score across all nodes in the path, ensuring that selected paths exhibit both high individual node quality and strong inter-level coherence. These paths serve directly as instruction tuning data, enabling MLLMs to learn explicit hierarchical reasoning patterns through instruction tuning.

%% file: image-and-table/overview.tex
\begin{figure}[!t]
    \centering
    \includegraphics[width=0.48\textwidth]{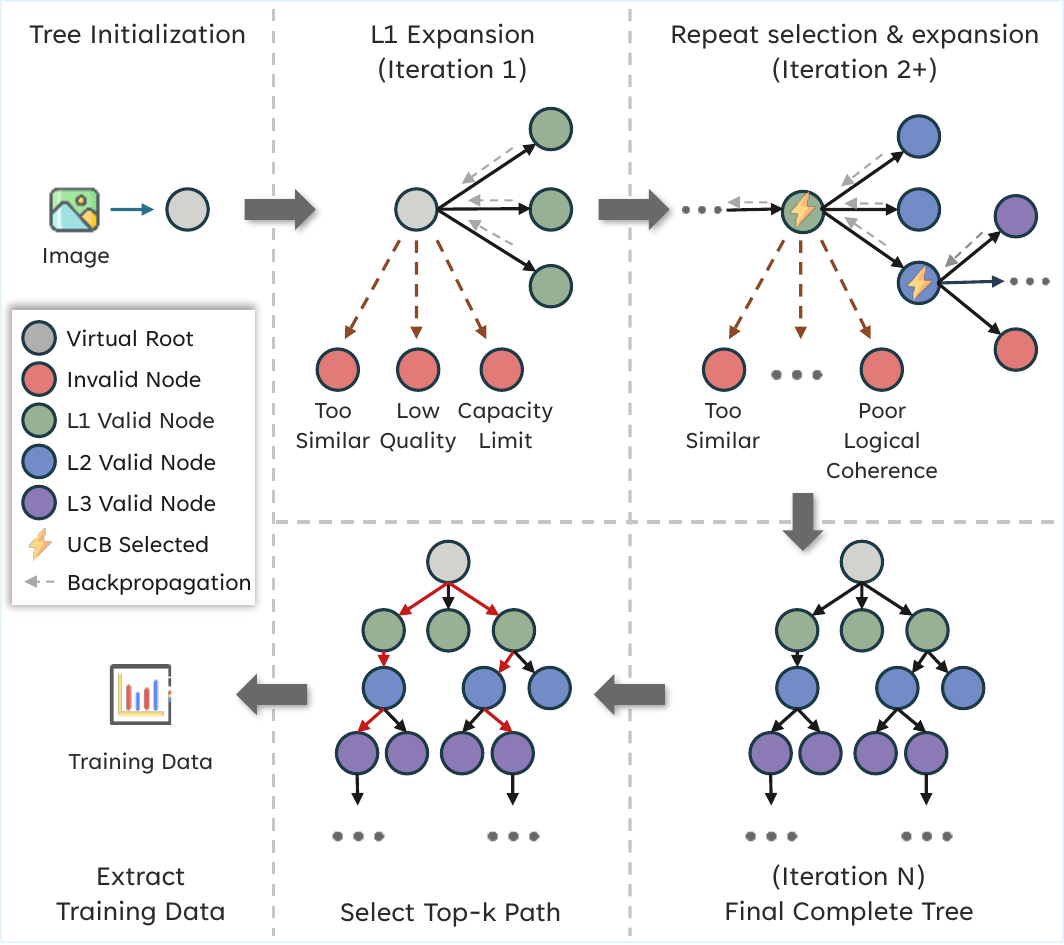}
    \caption{\textbf{Overview of our hierarchical data generation pipeline.} An MCTS-driven approach for generating high-quality hierarchical training data.}
    \label{fig:pipeline}
    \vspace{\myspace}
\end{figure}

%% file: image-and-table/main_results.tex
\begin{table*}[t]
\centering
\caption{\textbf{Overall results on \bench{}.} The best-performing model is \textbf{in-bold}, and the second is \underline{underlined}.}
\label{tab:multimodal_results}
\resizebox{\textwidth}{!}{
\begin{tabular}{l r cccc cccc cccc c}
    \toprule
    \multirow{2}{*}{\textbf{Model}} & \multirow{2}{*}{\textbf{Model Size}} & \multicolumn{4}{c}{\textbf{Implication Understanding}} & \multicolumn{4}{c}{\textbf{Aesthetic Appreciation}} & \multicolumn{4}{c}{\textbf{Affective Reasoning}} & \multirow{2}{*}{\textit{\textbf{Score}}} \\
    \cmidrule(lr){3-6} \cmidrule(lr){7-10} \cmidrule(lr){11-14}
     & & \bm{\aperc{}} & \bm{\abridge{}} & \bm{\aconn{}} & \bm{\afull{}} & \bm{\aperc{}} & \bm{\abridge{}} & \bm{\aconn{}} & \bm{\afull{}} & \bm{\aperc{}} & \bm{\abridge{}} & \bm{\aconn{}} & \bm{\afull{}} & \\
    \midrule
    \multicolumn{15}{l}{\textbf{Basic Reference}} \\
    \midrule
    Human & - & 99.25 & 96.00 & 86.50 & 86.00 & 99.14 & 92.29 & 90.29 & 88.86 & 99.33 & 93.33 & 88.67 & 86.67 & 87.18 \\
    GPT-4o & - & \textbf{95.50} & \underline{85.25} & \textbf{62.75} & \textbf{53.25} & \textbf{95.43} & \textbf{78.29} & \textbf{68.00} & \textbf{53.14} & 91.33 & 83.67 & \textbf{64.33} & \textbf{50.33} & \textbf{52.24} \\
    \midrule
    \multicolumn{15}{l}{\textbf{Open-Source MLLMs}} \\
    \midrule
    Qwen3-VL-Instruct & 4B & 86.75 & 82.75 & 58.00 & 43.25 & 90.57 & 70.86 & 60.00 & 41.14 & 90.33 & 82.67 & 56.67 & 39.33 & 41.24 \\
    Qwen3-VL-Instruct & 8B & 93.50 & \textbf{89.50} & 59.50 & \underline{50.75} & 91.71 & \underline{73.43} & \underline{63.43} & \underline{44.00} & \underline{94.33} & 84.67 & 60.00 & \underline{48.00} & \underline{47.58} \\
    LLaVA-1.6 & 7B & 81.75 & 58.00 & 40.25 & 18.75 & 79.14 & 36.86 & 33.14 & 9.43 & 92.00 & 58.00 & 19.33 & 12.00 & 13.39 \\
    LLaVA-1.6 & 13B & 84.75 & 79.00 & 55.00 & 39.50 & 84.86 & 55.14 & 50.57 & 26.29 & \underline{94.33} & 77.33 & 29.00 & 21.33 & 29.04 \\
    Deepseek-VL2-tiny & MoE 1B/3B & 88.25 & 62.25 & 49.75 & 29.25 & 89.71 & 45.14 & 41.14 & 19.71 & 93.33 & 65.33 & 29.00 & 19.00 & 22.65 \\
    Deepseek-VL2 & MoE 4.5B/27B & \underline{93.75} & 83.00 & 60.75 & 49.50 & \underline{95.14} & 58.00 & 38.00 & 23.71 & \textbf{96.33} & 81.33 & 46.00 & 36.67 & 36.63 \\
    Gemma3 & 4B & 76.50 & 72.00 & 49.75 & 30.75 & 68.86 & 62.86 & \textbf{68.00} & 29.14 & 87.00 & 76.00 & 51.00 & 36.00 & 31.96 \\
    Gemma3 & 12B & 87.50 & \underline{85.25} & 60.50 & 47.50 & 82.86 & 70.29 & \textbf{68.00} & 38.86 & 90.67 & \textbf{86.33} & 58.00 & 46.33 & 44.23 \\
    InternVL3.5 & 4B & 82.50 & 83.75 & 58.50 & 42.00 & 82.86 & 64.57 & 40.00 & 23.43 & 91.00 & 81.67 & \underline{60.67} & 47.67 & 37.70 \\
    InternVL3.5 & 8B & 82.00 & \underline{85.25} & 55.75 & 41.75 & 84.00 & 68.00 & 60.57 & 36.29 & 86.00 & 83.67 & 55.67 & 42.00 & 40.01 \\
    Phi-4-Multimodal-Instruct & 6B & 90.25 & 56.50 & 42.75 & 32.25 & 90.29 & 42.57 & 23.14 & 15.14 & 90.00 & \underline{85.00} & 45.33 & 33.67 & 27.02 \\
    Phi-3.5-Vision-Instruct & 4B & 84.25 & 83.25 & \underline{61.25} & 44.75 & 88.29 & 61.14 & 53.43 & 33.14 & 91.33 & 82.00 & 54.33 & 41.33 & 39.74 \\
    \bottomrule
\end{tabular}
}
\vspace{\myspace}
\end{table*}

%% file: sec/4_experiments.tex
\section{Experiments}

\subsection{Experimental Setup}

\paragraph{Models.}
To comprehensively assess \topic{} capabilities across diverse model architectures and scales, we evaluate 13 MLLMs on \bench{}. Our evaluation includes the leading proprietary model GPT-4o~\cite{hurst2024gpt}, alongside a broad spectrum of open-source models spanning multiple architectural families and parameter scales: Qwen3-VL-Instruct (4B, 8B)~\cite{qwen3technicalreport, Qwen2.5-VL, Qwen2VL, Qwen-VL}, LLaVA-1.6 (7B, 13B)~\cite{liu2024llavanext, liu2023improvedllava, liu2023llava}, Gemma3 (4B, 12B)~\cite{gemmateam2025gemma3technicalreport}, InternVL3.5 (4B, 8B)~\cite{wang2025internvl35advancingopensourcemultimodal}, Phi-4-Multimodal-Instruct (6B)~\cite{abdin2024phi4technicalreport}, and Phi-3.5-Vision-Instruct (4B)~\cite{abdin2024phi3technicalreporthighly}. We further include Mixture-of-Experts architectures with Deepseek-VL2-tiny (MoE 1B/3B) and Deepseek-VL2 (MoE 4.5B/27B)~\cite{wu2024deepseekvl2mixtureofexpertsvisionlanguagemodels} to examine whether sparse expert routing confers advantages for bridging perceptual and connotative levels. Finally, we evaluate \lora{}, our instruction-tuned model based on Qwen3-VL-4B-Instruct, to validate the efficacy of our hierarchical data generation pipeline. Complete training configurations and hyperparameters are detailed in Appendix~\ref{sec:implementation}. This diverse model selection enables systematic analysis of how architecture, scale, and hierarchical instruction data influence \topic{} performance.

\paragraph{Evaluation Benchmark.}
We evaluate all models on \bench{}, reporting results on its held-out test split. We conduct evaluations under two settings: a ``base'' setting where models answer each level independently, and a ``context'' setting where previous levels' predictions are provided as context. This design allows us to isolate and quantify the contribution of hierarchical dependencies. To assess whether instruction tuning on hierarchical data improves general reasoning capabilities beyond patterns specific to \bench{}, we evaluate \lora{} on four benchmarks: MMBench~\cite{liu2023mmbench}, MMMU~\cite{mmmu}, MMStar~\cite{mmstar}, and HallusionBench~\cite{guan2024hallusionbench}. These evaluations follow standard protocols without task-specific tuning.

\paragraph{Human Evaluation.}
To establish a human performance baseline, we collected evaluations from undergraduate students. For each \topic{} level (\lperc{}, \lbridge{}, and \lconn{}), we recruited three students to evaluate it and reported the average score for that level. We intentionally used different groups of annotators for each level to prevent information leakage between them. This ensures the evaluation mirrors the ``base" setting for models, as using the same person to evaluate all levels would inadvertently replicate the ``context" setting.

\subsection{Main Results}

\paragraph{Gap between Humans and MLLMs}
As shown in Table~\ref{tab:multimodal_results}, humans nearly saturate all \bench{} levels across all tasks, achieving an overall score of 87.18\% with consistently strong performance at both \lbridge{} and \lconn{}. In contrast, top models fall far behind at higher abstraction levels. On Implication Understanding, humans achieve 99.25\% at \lperc{}, 96.00\% at \lbridge{}, and 86.50\% at \lconn{}, while GPT-4o demonstrates near-human performance at \lperc{} (-3.75\%) but exhibits a substantial disparity at \lconn{} (-23.75\%). The strongest open-source baseline, Qwen3-VL-8B-Instruct, demonstrates a similar pattern with performance gaps of -5.75\% at \lperc{} and -27.00\% at \lconn{}. These level-specific disparities result in significantly lower \afull{}, with GPT-4o and Qwen3-VL-8B-Instruct underperforming humans by -32.75\% and -35.25\% respectively. This pattern holds consistently across all three tasks, with the overall score revealing a substantial gap of -34.94\% for GPT-4o and -39.60\% for Qwen3-VL-8B-Instruct, demonstrating that current MLLMs still lack a stable semantic bridge from concrete evidence to abstract meaning.

\paragraph{Universal Performance Degradation in \topic{}.}
Beyond this human-model gap, Table~\ref{tab:multimodal_results} reveals a widespread degradation pattern across evaluated MLLMs. Most models, regardless of scale or architecture, exhibit a sharp, cascading decline when moving from perception to connotation. While nearly all achieve high accuracy at \lperc{}, their performance systematically deteriorates at \lbridge{} and typically drops precipitously at \lconn{}. This pattern is pronounced in strong models: GPT-4o experiences a degradation of -32.75\%, Qwen3-VL-8B-Instruct degrades by -34.00\%, and Gemma3-12B suffers a -27.00\% decline on Implication Understanding. Even models that deviate from this trend on specific tasks still struggle at \lconn{}, leading to consistently low \afull{}. Such cascading error propagation validates that a critical weakness exists in bridging perception to abstract reasoning.

\input{image-and-table/context.tex}
\paragraph{Analysis of Model Scale and Architecture.}
Table~\ref{tab:multimodal_results} further reveals that while increasing model scale generally improves performance, it does not resolve the fundamental challenges of \topic{}. For example, LLaVA-1.6-13B achieves an overall score of 29.04\%, significantly outperforming its 7B counterpart at 13.39\%, yet its performance at \lconn{} remains weak. This suggests that larger models possess stronger foundational capabilities but still lack the specialized knowledge required for connotative reasoning. We also observe distinct performance profiles across different model architectures. Certain architectures, such as Qwen3-VL exhibit more balanced performance across levels, whereas LLaVA-1.6 shows a particularly steep decline after \lperc{}, indicating a weaker ability to understand the semantic bridge. These patterns strongly suggest that the challenge of \topic{} transcends mere model scale and architecture, pointing instead to a deeper, more fundamental gap in current multimodal understanding paradigms.

\paragraph{Hierarchical Dependency in \bench{}.}
To verify that performance on higher levels depends on lower-level capabilities, we conducted a comparative analysis between the ``base'' and ``context'' settings. As shown in Table~\ref{tab:context_ablation} (detailed results in Appendix~\ref{subsec:context_detailed_results}), providing hierarchical context yields substantial performance gains across all evaluated models. Specifically, GPT-4o demonstrates an overall improvement of +15.94\%, while Qwen3-VL-8B-Instruct achieves a gain of +14.70\%. Examining the level-wise improvements, we observe that most models, particularly stronger ones, exhibit marked accuracy gains across \lbridge{} and \lconn{}, with GPT-4o improving by approximately +3.6\% and +16.6\% on average across the two levels, respectively. Even in cases where certain tasks show irregular patterns, such as Gemma3-4B on Affective Reasoning, the overall score consistently improves (+4.43\%), a pattern that further aligns with our expectation that models with weaker lower-level capabilities would show more constrained higher-level improvements. These pervasive and significant uplifts demonstrate that \bench{} indeed exhibits strong inter-level dependencies, with lower levels providing critical grounding for higher-level connotative reasoning. Beyond validating our hierarchical evaluation, this finding underscores that connotative inference is not a monolithic task but a hierarchical process that fundamentally relies on a coherent chain of reasoning from perception through semantic bridging to abstract interpretation.

\subsection{Effectiveness of Hierarchical Data Generation}

\input{image-and-table/data_augmentation.tex}

To validate the effectiveness of our data generation pipeline, we instruction-tune Qwen3-VL-4B-Instruct on approximately 10k hierarchical QA pairs generated from 1k images (Section~\ref{subsec:data_generation}), yielding \lora{}.

\paragraph{Improvements and Generalization.}
As shown in Figure~\ref{fig:lora}~(left), \lora{} exhibits consistent improvements across all three tasks of \bench{}. Although training supervision is provided exclusively for Implication Understanding, the model achieves substantial gains not only on this task (+6.75\% in \afull{}) but also on Aesthetic Appreciation (+5.43\%) and Affective Reasoning (+6.34\%), where no direct training signals are given. This cross-task transfer strongly indicates that the model has learned to establish generalizable semantic connections, linking perceptual evidence to abstract connotations through intermediate factual reasoning, rather than memorizing task-specific templates. The overall \bench{} score demonstrates an improvement of +6.17\%, further confirming that bottom-up data generation with validation effectively teaches structured visual reasoning. Critically, our data generation pipeline operates independently of \bench{}'s specific task formulations and answer patterns, relying solely on generic hierarchical reasoning prompts. This ensures that the observed gains reflect genuine improvements in hierarchical inference rather than dataset-specific overfitting (detailed results in Appendix~\ref{subsec:lora_detailed_results}).

To assess whether these hierarchical reasoning improvements generalize beyond \bench{}, we evaluate \lora{} on four established general benchmarks: MMBench, HallusionBench, MMStar, and MMMU. As shown in Figure~\ref{fig:lora}~(right), \lora{} demonstrates strong generalization, achieving substantial improvements on MMStar (+7.26\%) and MMMU (+3.22\%) while maintaining competitive performance on MMBench and HallusionBench, even though it was trained exclusively on \bench{}'s connotation-focused data (see Appendix~\ref{subsec:lora_detailed_results} for complete results). This cross-benchmark transfer demonstrates that grounding abstract interpretations in perceptual evidence through a structured semantic bridge enhances reasoning skills, benefiting tasks beyond connotative understanding.

\input{image-and-table/lowleveluseful}

\paragraph{Impact of Hierarchical Training Data.}
We compare three training setups in Table~\ref{tab:lowleveluseful} to examine how hierarchical supervision affects model learning. \textbf{Full-data} uses complete three-level supervision across \lperc{}, \lbridge{}, and \lconn{}, yielding approximately 30k QA pairs as described in Appendix~\ref{sec:implementation}. \textbf{L3-only} trains exclusively on \lconn{} pairs, producing 10k examples that span all data entries in Full-data but only at the top reasoning level. \textbf{Full-hierarchy} samples a balanced 10k subset from Full-data by selecting one-third of its complete hierarchical entries, where each retained entry includes all three levels for that specific reasoning chain. This ensures Full-hierarchy matches L3-only's count while preserving multi-level structure, but exposes the model to fewer reasoning chains than L3-only.

Results demonstrate the effectiveness of hierarchical supervision. Full-hierarchy outperforms L3-only by +1.25\% despite training on fewer complete entries. This confirms that \lperc{} and \lbridge{} supervision provide critical learning signals that explicitly teach the model how to ground abstract connotations in perceptual evidence through structured intermediate steps. Full-data further improves performance by +1.75\%, showing that scaling hierarchical supervision amplifies these benefits.

\subsection{Case Study}

\input{image-and-table/case_study}

Figure~\ref{fig:case_study} demonstrates how training on hierarchical data corrects semantic bridging failures. In this fly swatter display example, the base model Qwen3-VL-4B-Instruct correctly identifies the object at \lperc{} but fails at \lbridge{}, misinterpreting the display as visual contrast rather than recognizing it as a trophy-room parody, which propagates to incorrect interpretation at \lconn{}. In contrast, \lora{} successfully constructs the complete reasoning chain by establishing the correct semantic bridge, demonstrating that our hierarchical data explicitly teaches models to ground abstract connotations through structured intermediate reasoning. This directly addresses the \lbridge{} bottleneck, where the improvement reflects genuine learning of systematic connections between perceptual evidence and abstract meanings rather than task-specific memorization.

%% file: image-and-table/context.tex
\begin{table}[t!]
\centering
\caption{\textbf{Comparison of ``base'' and ``context'' settings on \bench{}.} $+\Delta$ denotes gain over ``base''.}
\label{tab:context_ablation}
\resizebox{\columnwidth}{!}{%
\begin{tabular}{l l cccc c}
\toprule
\multirow{2}{*}{\textbf{Model}} & \multirow{2}{*}{\textbf{Task}} & \multicolumn{4}{c}{\textbf{Accuracy}} & \multirow{2}{*}{\textit{\textbf{Score}}} \\
\cmidrule(lr){3-6}
& & \bm{\aperc{}} & \bm{\abridge{}} & \bm{\aconn{}} & \bm{\afull{}} & \\
\midrule
\multirow{3}{*}{GPT-4o(Base)} & Impl. & 95.50 & 85.25 & 62.75 & 53.25 & \multirow{3}{*}{52.24} \\
& Aesth. & 95.43 & 78.29 & 68.00 & 53.14 & \\
& Affect. & 91.33 & 83.67 & 64.33 & 50.33 & \\
\addlinespace[1pt]
\hdashline[2pt/2pt]
\addlinespace[1pt]
\multirow{3}{*}{GPT-4o(Context)} & Impl. & 95.50 & 89.75 & 76.50 & 65.00 & \multirow{3}{*}{\makecell{68.18 \\ \colorbox{gray!15}{+15.94}}} \\
& Aesth. & 95.43 & 82.29 & 87.71 & 72.86 & \\
& Affect. & 91.33 & 86.00 & 80.67 & 66.67 & \\
\addlinespace[1pt]
\midrule
\addlinespace[1pt]
\multirow{3}{*}{Qwen3-VL-8B(Base)} & Impl. & 93.50 & 89.50 & 59.50 & 50.75 & \multirow{3}{*}{47.58} \\
& Aesth. & 91.71 & 73.43 & 63.43 & 44.00 & \\
& Affect. & 94.33 & 84.67 & 60.00 & 48.00 & \\
\addlinespace[1pt]
\hdashline[2pt/2pt]
\addlinespace[1pt]
\multirow{3}{*}{Qwen3-VL-8B(Context)} & Impl. & 93.50 & 90.00 & 74.75 & 62.75 & \multirow{3}{*}{\makecell{62.28 \\ \colorbox{gray!15}{+14.70}}} \\
& Aesth. & 91.71 & 74.00 & 82.57 & 59.43 & \\
& Affect. & 94.33 & 89.00 & 76.00 & 64.67 & \\
\addlinespace[1pt]
\midrule
\addlinespace[1pt]
\multirow{3}{*}{Gemma3-4B(Base)} & Impl. & 76.50 & 72.00 & 49.75 & 30.75 & \multirow{3}{*}{31.96} \\
& Aesth. & 68.86 & 62.86 & 68.00 & 29.14 & \\
& Affect. & 87.00 & 76.00 & 51.00 & 36.00 & \\
\addlinespace[1pt]
\hdashline[2pt/2pt]
\addlinespace[1pt]
\multirow{3}{*}{Gemma3-4B(Context)} & Impl. & 76.50 & 78.25 & 63.50 & 40.75 & \multirow{3}{*}{\makecell{36.39 \\ \colorbox{gray!15}{+4.43}}} \\
& Aesth. & 68.86 & 65.14 & 82.57 & 35.43 & \\
& Affect. & 87.00 & 75.00 & 50.00 & 33.00 & \\
\bottomrule
\end{tabular}
}
\vspace{\myspace}
\end{table}

%% file: image-and-table/data_augmentation.tex
\begin{figure}[!t]
    \centering
    \includegraphics[width=0.48\textwidth]{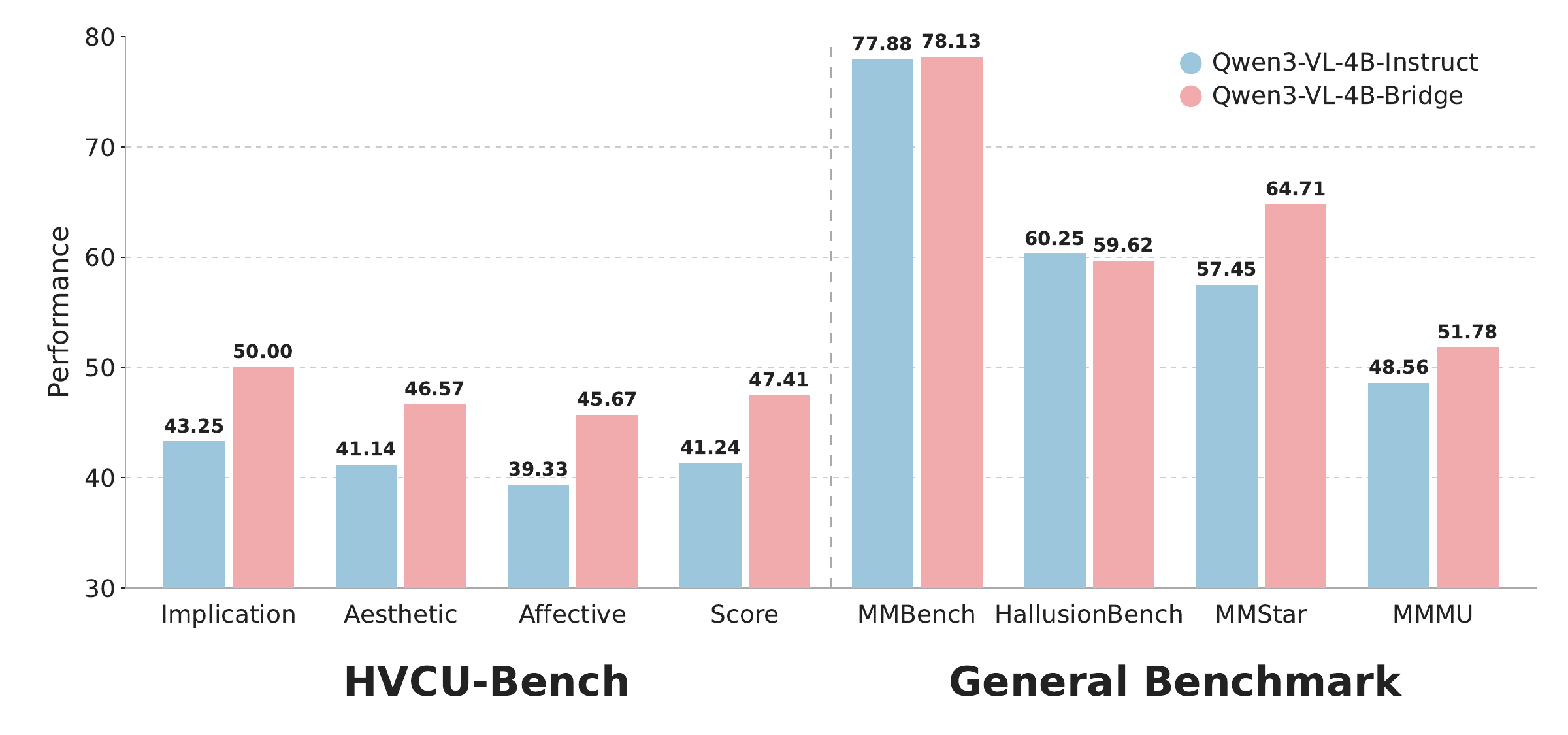}
    \caption{\textbf{Performance of \lora{}} on \textbf{(left)} \bench{} and \textbf{(right)} general benchmarks.}
    \label{fig:lora}
    \vspace{-4mm}
\end{figure}

%% file: image-and-table/lowleveluseful.tex
\begin{table}[t!]
\centering
\caption{\textbf{Training data composition comparison.} $+\Delta$ denotes gain over L3-only.}
\label{tab:lowleveluseful}
\resizebox{\columnwidth}{!}{%
\begin{tabular}{lrcccc}
\toprule
\textbf{Data} & \textbf{\#QA} & \bm{\aperc{}} & \bm{\abridge{}} & \bm{\aconn{}} & \bm{\afull} \\
\midrule
L3-only & 10k & 89.25 & 85.75 & 59.50 & 47.00 \\
\addlinespace[1pt]
\hdashline[2pt/2pt]
\addlinespace[1pt]
Full-hierarchy & 10k & 89.75\tiny{\colorbox{gray!10}{$+0.50$}} & 86.25\tiny{\colorbox{gray!10}{$+0.50$}} & 60.75\tiny{\colorbox{gray!10}{$+1.25$}} & 48.25\tiny{\colorbox{gray!10}{$+1.25$}} \\
Full-data & 30k & 93.50\tiny{\colorbox{gray!10}{$+4.25$}} & 89.25\tiny{\colorbox{gray!10}{$+3.50$}} & 63.75\tiny{\colorbox{gray!10}{$+4.25$}} & 50.00\tiny{\colorbox{gray!10}{$+3.00$}} \\
\bottomrule
\end{tabular}
}
\vspace{\myspace}
\end{table}

%% file: image-and-table/case_study.tex
\begin{figure}[!t]
    \centering
    \includegraphics[width=0.48\textwidth]{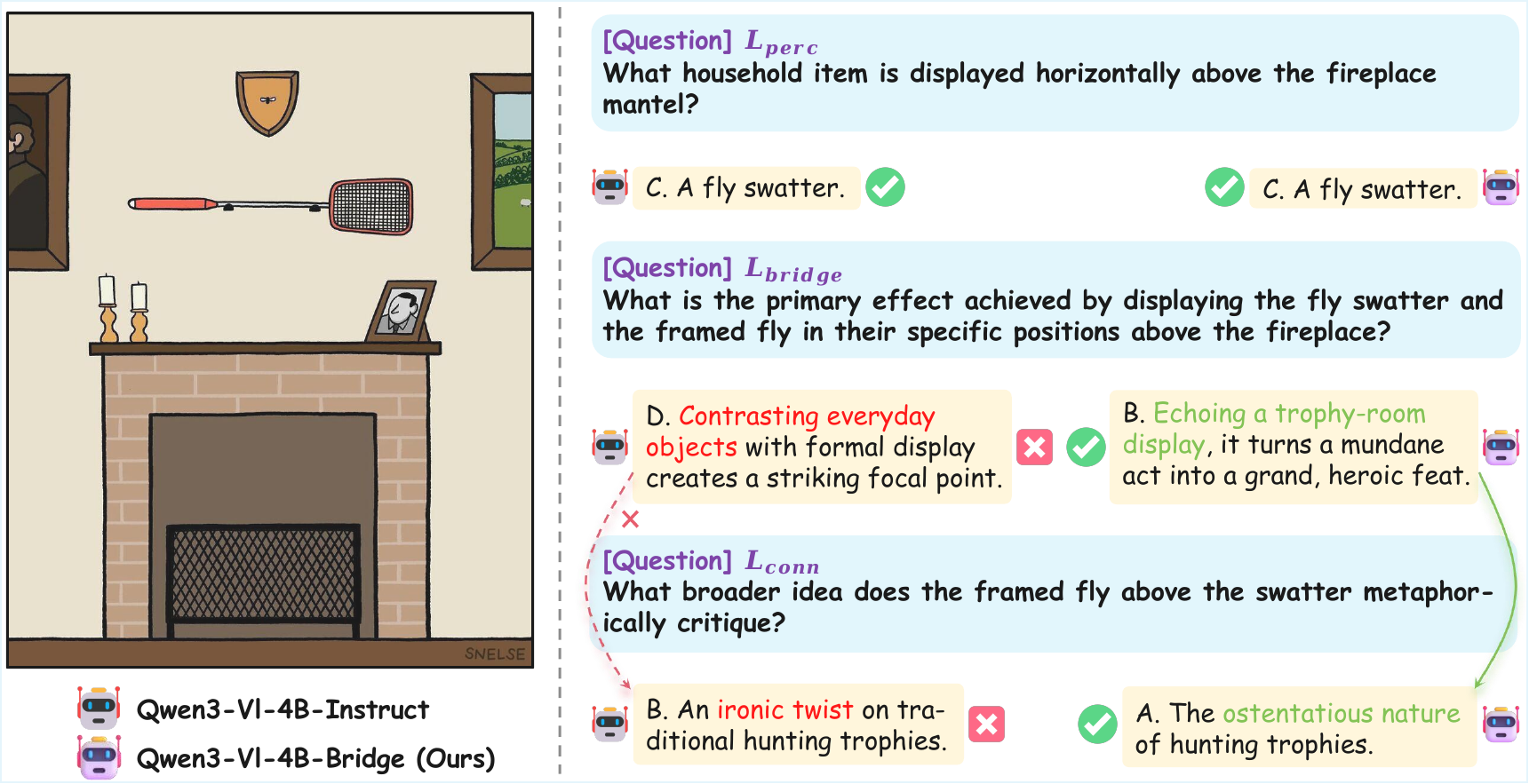}
    \caption{\textbf{Representative case study.} Training on hierarchical data corrects semantic bridging failures.}
    \label{fig:case_study}
    \vspace{\myspace}
\end{figure}

%% file: sec/5_Conclusion.tex
\section{Conclusion}

This paper addressed the challenge of bridging visual perception and abstract interpretation in MLLMs. We introduced \topic{}, a hierarchical framework progressing from perception through semantic bridging to connotation, explicitly modeling the inferential process that connects concrete visual evidence to abstract meanings. Through \bench{}, we revealed that all evaluated models exhibit universal performance degradation across levels, with semantic bridging emerging as the critical bottleneck, while hierarchical context experiments demonstrated that lower-level perceptual foundations provide crucial grounding for higher-level interpretation. To address this gap, our MCTS-driven data generation pipeline produces hierarchical data for instruction tuning, yielding substantial improvements on both \bench{} and general benchmarks, demonstrating that explicit hierarchical supervision is essential for teaching models human-like visual understanding.

%% file: sec/X_suppl.tex
\clearpage
\appendix
\onecolumn
\setcounter{page}{1}
\maketitlesupplementary

\section{More Related Works}
\label{sec:more_related_works}

While instruction data for MLLMs has evolved toward automated synthesis, existing pipelines rarely produce explicit, verifiable chains connecting perception to connotation. Most approaches prioritize data scale or quality over intermediate reasoning steps that bridge visual observations with abstract interpretations.

\paragraph{Template-based Generation.} 
Pipelines such as LLaVA-Instruct-150K~\cite{liu2023visualinstructiontuning}, InstructBLIP~\cite{dai2023instructblip}, and ShareGPT4V~\cite{chen2023sharegpt4v} leverage powerful language models to generate large-scale instruction-following data from image captions or dense descriptions. While these methods successfully scale data coverage, they primarily produce \textit{flat} question-answer pairs that lack hierarchical structure. The generated data provides weak supervision for intermediate reasoning steps, leaving the inferential connections between visual evidence and abstract interpretations underspecified. Consequently, models trained on such data learn to map images directly to high-level conclusions without explicitly modeling the semantic bridge that justifies these interpretations.

\paragraph{Search-based Generation.} 
Recent work has begun integrating tree search mechanisms to explore diverse reasoning paths. Socratic-MCTS~\cite{acuna2025socraticmctstesttimevisualreasoning} applies MCTS-style procedures at inference time to elicit subquestions and intermediate verification steps, improving interpretability without additional training. ReST-MCTS*~\cite{zhang2024restmctsllmselftrainingprocess} employs MCTS with process-level rewards to curate high-quality reasoning trajectories in language modeling. While these methods demonstrate the potential of search-driven data generation, they lack \textit{structured hierarchical progression} from concrete perceptual evidence to abstract connotative meaning. Moreover, they typically do not enforce \textit{inter-level validation} to ensure that each reasoning step logically supports the next, which is critical for learning coherent semantic bridges. In contrast, our MCTS-driven pipeline explicitly constructs complete hierarchical reasoning chains with bottom-up search and inter-level validation, ensuring that generated data teaches models to ground abstract meanings in concrete visual evidence through verifiable intermediate steps.

\section{\bench{} Details}
\label{sec:bench_details}
\subsection{Dataset Statistics}
As shown in Table \ref{tab:statistics}, \bench{} comprises 1,050 hierarchical multiple-choice QA samples distributed across three task families (Implication Understanding, Aesthetic Appreciation, and Affective Reasoning) and fifteen fine-grained aspects. Each sample is organized as a three-level question–answer chain (\lperc{}, \lbridge{}, \lconn{}) defined on a single image, resulting in a total of 3,150 QA pairs. This design supports both level-wise diagnosis and full-chain evaluation of \topic{}.

\input{image-and-table/statistics}
\subsection{Data Quality Assurance and Validation}
To further ensure the quality and reliability of \bench{}, we perform a multi-stage human auditing process on top of the model-driven generation framework described in Section \ref{subsec:bench}. 

\paragraph{Validation Protocol.}
After automatic format and consistency checks, every candidate hierarchical multiple-choice QA chain is manually inspected by human annotators with access to the original image and all three \topic{} levels. Each question–answer pair is validated by at least five annotators, who independently verify that (i) the correct option at each level is unambiguously supported by the visual content, and (ii) the three levels together form a logically coherent reasoning chain from perception to connotation.

\paragraph{Consensus Resolution.}
We monitor inter-annotator agreement and treat any inconsistencies or potential misinterpretations as flags for further review. Disagreements are resolved through discussion within the annotation team until a single consensus label and reasoning chain are reached; items that remain ambiguous or exhibit artifacts from the generation model are discarded rather than forced to consensus.

\paragraph{Multi-Round Quality Control.}
In total, we conduct three rounds of such quality control, including repeated passes of cross-checking and consolidation, before finalizing the benchmark. As a result, the released \bench{} uses human-validated annotations rather than raw outputs from Gemini-2.5-Pro, substantially reducing the risk that systematic biases or errors from the generation model propagate into our evaluation data.

\subsection{Dataset Samples}
\label{appendix:dataset_samples}
To provide concrete illustrations of the hierarchical reasoning structure in \bench{}, we present representative samples spanning all three task families and fifteen fine-grained aspects. Each sample demonstrates a three-level question–answer chain from perception to connotation, illustrating how the benchmark grounds abstract interpretations in concrete visual evidence through explicit semantic bridges.

\paragraph{List of Samples}

\begin{enumerate}
  \item \textit{Implication Understanding}
  \begin{enumerate}
    \item \textit{Metaphor} \dotfill \pageref{fig:implication_metaphor}
    \item \textit{Contrast} \dotfill \pageref{fig:implication_unicorn_case}
    \item \textit{Exaggeration} \dotfill \pageref{fig:test108}
    \item \textit{Dislocation} \dotfill \pageref{fig:test1182}
    \item \textit{Symbolism} \dotfill \pageref{fig:test873}
  \end{enumerate}
  \item \textit{Affective Reasoning}
  \begin{enumerate}
    \item \textit{Fear} \dotfill \pageref{fig:test122}
    \item \textit{Joy} \dotfill \pageref{fig:test9}
    \item \textit{Wonder} \dotfill \pageref{fig:test270}
    \item \textit{Anger} \dotfill \pageref{fig:test257}
    \item \textit{Affection} \dotfill \pageref{fig:test117}
    \item \textit{Sadness} \dotfill \pageref{fig:test30}
  \end{enumerate}
  \item \textit{Aesthetic Appreciation}
  \begin{enumerate}
    \item \textit{Graphic} \dotfill \pageref{fig:graphic240}
    \item \textit{Color} \dotfill \pageref{fig:color164}
    \item \textit{Font} \dotfill \pageref{fig:font27}
    \item \textit{Composition} \dotfill \pageref{fig:composition210}
  \end{enumerate}
\end{enumerate}

\section{Implementation Details}
\label{sec:implementation}
\paragraph{Data Generation.}
We generate approximately 10k hierarchical instruction data using Gemini 2.5-Pro and our MCTS-driven pipeline (Section~\ref{subsec:data_generation}) from 1k images. For each image, MCTS explores a 3-level reasoning tree with exploration constant $c=2.0$, expanding up to 5 candidate nodes per step with quality filtering (threshold 0.65) and diversity control (threshold 0.75). Tree capacity is limited to 8, 12, and 15 nodes per level. We extract the top-10 paths per image, yielding approximately 10k three-level hierarchical chains, totaling around 30k question-answer pairs. All data follows an open-ended QA format rather than multiple-choice, ensuring the model learns genuine hierarchical reasoning instead of selection patterns.

\paragraph{Model Training.}
Our \lora{} is instruction-tuned from Qwen3-VL-4B-Instruct on this data using LoRA with rank 32 and $\alpha=64$. Training is performed for 3 epochs with a learning rate of $2.0 \times 10^{-5}$, batch size of 128, and AdamW optimizer, implemented using the LLaMA-Factory~\cite{zheng-etal-2024-llamafactory}. All experiments are conducted on NVIDIA A100 GPUs with 80GB of memory, requiring approximately 8 GPU-hours.

\paragraph{Evaluation Protocol.}
For \bench{} evaluation, models are prompted to select from four options (A/B/C/D) for each multiple-choice question in a zero-shot setting. All models are evaluated with the temperature set to 0 to ensure deterministic outputs. We use a rule-based parser to extract answer choices from model outputs, treating unparseable or invalid responses as incorrect. Open-source models are evaluated locally with their default inference configurations, while proprietary models (e.g., GPT-4o) are accessed via official APIs with temperature=0.

\section{More Experimental Details}
\label{sec:more_exp_details}
\input{image-and-table/lora_results}
\subsection{Detailed Results of \lora{}}
\label{subsec:lora_detailed_results}
\paragraph{Evaluation on \bench{}.}
Table \ref{tab:lora_results} provides the complete results corresponding to Figure \ref{fig:lora}~(left), reporting the \bench{} results of \lora{} under both the ``base'' and ``context'' settings. In the ``base'' setting, \lora{} achieves strong perceptual accuracy (\aperc{} $>$ 90\% across all tasks) while substantially improving \lbridge{}, \lconn{}, and \afull{}, reaching an overall score of 47.41\%. The ``context'' setting reveals even stronger performance: \lora{} achieves 58.83\%, outperforming the base model Qwen3-VL-4B-Instruct by +3.73\% (55.10\% in Table \ref{tab:detailed_results}). When transitioning from ``base'' to ``context'', \lora{} shows substantial gains in full-chain accuracy, improving by +8.25\%, +12.00\%, and +14.00\% on Implication Understanding, Aesthetic Appreciation, and Affective Reasoning, respectively. These results demonstrate that \topic{} effectively teaches hierarchical reasoning chains that maintain superior performance across both evaluation settings, confirming that our training approach enhances the model's intrinsic reasoning capabilities rather than merely optimizing for specific evaluation conditions.

\paragraph{Evaluation on General Benchmarks.}
Table \ref{tab:general_benchmarks} provides the complete results corresponding to Figure \ref{fig:lora}~(right), comparing Qwen3-VL-4B-Instruct and \lora{} on four general benchmarks: MMBench, HallusionBench, MMStar, and MMMU. We evaluate using the LMMS-Eval~\cite{zhang2025lmms} toolkit and report the official metrics defined by each benchmark. For benchmarks with multiple sub-metrics (MMBench and MMStar), we also report the average score across all sub-metrics for easier comparison. \lora{} maintains strong performance on MMBench and HallusionBench while achieving substantial gains on MMStar (Avg. +7.26\%) and MMMU (+3.22\%). Aggregated across all four benchmarks, \lora{} achieves an average improvement of +2.53\%. These results confirm that the hierarchical training data does not cause overfitting to \bench{}, but instead induces hierarchical reasoning patterns that transfer to diverse external benchmarks.
\input{image-and-table/general_benchmarks}

\subsection{``Context'' Setting of \bench{}}
\label{subsec:context_detailed_results}
Table \ref{tab:detailed_results} presents the full \bench{} results under the ``context'' setting, which can be compared against the ``base'' setting results in Table \ref{tab:multimodal_results}. In the ``context'' setting, models are given predictions from preceding levels as additional context, allowing us to explicitly probe whether access to lower-level information facilitates higher-level connotative reasoning.

\paragraph{Universal Improvement Across Models.}
The results reveal a consistent trend: when provided with hierarchical context, almost every model exhibits substantial improvements across \abridge{}, \aconn{}, and \afull{}. This phenomenon spans the entire spectrum of model capabilities, from weaker architectures like LLaVA-1.6-7B (+9.85\%) to stronger models like GPT-4o (+15.94\%). This universal improvement provides robust empirical evidence for strong inter-level dependency, confirming that connotative reasoning is causally linked to the availability and quality of foundational information rather than being an isolated capability that could operate independently of hierarchical context.

\paragraph{Model-Specific Bottlenecks.}
The magnitude of these gains reveals different limitations across models. Stronger models such as GPT-4o and Qwen3-VL-8B achieve the largest absolute improvements (+15.94\% and +14.70\% respectively), suggesting that their primary bottleneck in the ``base'' setting is missing spontaneous connectivity rather than insufficient reasoning capacity. In contrast, weaker models also improve, confirming they follow the same hierarchical logic, but their performance ceiling remains constrained by the absolute quality of their underlying perception (\aperc{}), a limitation that even perfect reasoning cannot fully offset. Together, these differential patterns reveal that hierarchical visual understanding requires both robust perception and effective cross-level connectivity.
\input{image-and-table/detailed_results}

\subsection{Application on Larger and Heterogeneous Models}
To assess the scalability and architectural universality of our approach, Table \ref{tab:other_model} presents the impact of hierarchical instruction tuning on a larger model (Qwen3-VL-8B) and a heterogeneous architecture (LLaVA-1.6-7B).

\paragraph{Scalability to Larger Models.}
We instruction-tune Qwen3-VL-8B-Instruct on the same data to obtain Qwen3-VL-8B-Bridge. The results reveal a striking pattern: while the 8B model already achieves near-saturated perceptual accuracy (\aperc{} $>$ 93\% across all tasks), hierarchical tuning yields substantial gains at higher reasoning levels, with \afull{} improving by +4.25\%, +8.57\%, and +1.00\% on the three task families. This asymmetric improvement profile suggests that larger models already possess strong capabilities at individual levels, but lack the explicit pathways connecting them. Our hierarchical data acts as a structural scaffold that bridges the model's latent low-level perceptual knowledge with its high-level reasoning capacity, enabling cross-level integration. Rather than teaching new perceptual skills to an already-saturated foundation, the training data unlocks the model's ability to systematically propagate visual evidence through intermediate semantic reasoning to abstract interpretations, transforming isolated competencies into a coherent reasoning chain.

\paragraph{Architectural Robustness.}
To verify architectural generality, we apply the same data to LLaVA-1.6-7B, which employs a fundamentally different vision-language fusion mechanism. The model achieves substantial improvements in \afull{} (+14.50\% on Implication Understanding, +6.28\% on Aesthetic Appreciation), with an overall score gain of +7.60\%. Notably, the Aesthetic Appreciation task exhibits a slight \aperc{} regression (-3.43\%), likely reflecting LLaVA-1.6's relatively limited baseline perceptual capacity on this specific task. However, this minor drop at the perceptual level does not prevent significant gains at \abridge{} and \aconn{}, which is particularly revealing: it demonstrates that hierarchical supervision can successfully teach higher-level reasoning even when lower-level perception remains imperfect or unstable. In other words, the effectiveness of our training framework does not depend on a flawless perceptual foundation, but rather on establishing coherent cross-level reasoning pathways that can function robustly across diverse architectural designs and varying baseline capabilities, confirming the architectural universality of our hierarchical supervision approach.
\input{image-and-table/other_model}

\subsection{Comparison with Other Data Generation Approaches}
To validate the effectiveness of our MCTS-driven hierarchical generation, we compare it against a direct generation baseline that produces three-level questions without inter-level validation. Both generate data from the same 1k Implication Understanding images, while Aesthetic Appreciation and Affective Reasoning use entirely different image distributions, enabling cross-distribution generalization assessment. Table \ref{tab:other_data} presents the results.

\paragraph{Direct Generation Baseline.}
We construct Qwen3-VL-4B-Direct by instruction-tuning the base model on data generated via a straightforward three-level prompting strategy. This baseline prompt instructs the MLLM to create questions at three difficulty levels (basic perception, connection \& relationship, high-level reasoning) around the same topic, but does not enforce bottom-up construction or validate that each level's reasoning logically supports the next. While this approach produces superficially similar hierarchical structures, the resulting questions exhibit weak inter-level dependency, with each level potentially addressing independent aspects of the image rather than forming a coherent reasoning chain.

\paragraph{Performance Comparison.}
Qwen3-VL-4B-Bridge achieves a +2.05\% overall score improvement over Qwen3-VL-4B-Direct. The performance pattern reveals two complementary strengths. First, on Implication Understanding, where training data originates from the same task distribution, Qwen3-VL-4B-Bridge demonstrates stronger full-chain reasoning (+1.25\% in \afull{}), confirming that inter-level validation produces more coherent reasoning chains. More strikingly, the advantages amplify on tasks from different distributions: Aesthetic Appreciation and Affective Reasoning exhibit substantially larger gains at both \aconn{} (+2.00\% and +4.33\%) and \afull{} (+2.57\% and +2.34\%). This cross-distribution generalization directly reflects the structural quality of our training data. The MCTS tree search explores diverse reasoning paths and selects top-K chains, producing training examples with greater structural variety than direct generation. Rather than overfitting to specific task characteristics, the model learns generalizable hierarchical reasoning strategies that transfer robustly across different visual and semantic distributions, confirming that our approach produces training data with both strong inter-level coherence and broad applicability across diverse reasoning tasks.
\input{image-and-table/data_generation_comparision}

\section{Limitations}
\paragraph{Limitations of Three-Level Hierarchy.}
Although \bench{} and \topic{} provide a first step toward systematically modeling visual connotative hierarchical understanding, several limitations remain. First, our formulation currently instantiates \topic{} as a three-level discrete hierarchy over three task families (Implication Understanding, Aesthetic Appreciation, and Affective Reasoning). While this design captures a broad range of connotative phenomena, it is still a simplified approximation of human visual cognition, which may involve more continuous, overlapping, or multi-path reasoning processes. In particular, some real-world connotation and aesthetic judgments may not decompose cleanly into a single canonical chain from \lperc{} to \lconn{}, and our benchmark cannot fully represent such richer structures.

\paragraph{Scope of Task Coverage.} 
While \bench{} currently encompasses three major task families (Implication Understanding, Aesthetic Appreciation, and Affective Reasoning), the spectrum of visual connotation extends beyond these categories. For instance, interpreting complex cultural allusions or inferring subtle social dynamics involving multiple characters often requires specialized knowledge that goes beyond the current scope. Additionally, our benchmark focuses on static images, whereas narrative connotations in sequential images or video remain an unexplored frontier. Future work could expand the VCU-Bridge framework to encompass these broader domains, investigating whether the same hierarchical bridging mechanisms apply to these more dynamic forms of visual reasoning.

\paragraph{Cultural and Subjective Variability.}
Visual connotation is inherently subjective and culturally situated. To maximize generality, we employ a rigorous multi-stage human validation process with annotators from diverse cultural backgrounds to ensure consensus, and deliberately select images and design questions emphasizing universal connotative patterns. However, despite these efforts, the interpretations in \bench{} inevitably reflect certain cultural perspectives of the annotators and source data. Some symbolic meanings or aesthetic judgments may not generalize across all cultural contexts. Consequently, while our evaluation treats connotation as having a single ``ground truth" for standardized benchmarking, this may not fully capture the diverse nature of human interpretation across cultures, which we acknowledge as a direction for future research.

\section{Ethical Considerations} 
\paragraph{Copyright and Licensing.}
It is essential to strictly follow all copyright and licensing regulations. All images in \bench{} are sourced from publicly available datasets with appropriate research licenses. Data from sources that do not permit copying or redistribution will be explicitly excluded.

\paragraph{Data Privacy.}
Adherence to privacy laws and ethical standards in data handling is crucial. Annotators are explicitly instructed to avoid selecting images or creating questions that contain personally identifiable information. All selected images undergo privacy review before inclusion in the dataset.

\section{Prompts}
\label{sec:prompts}
We present the complete set of prompts used in both the \bench{} benchmark construction pipeline and the hierarchical instruction data generation pipeline. The prompts are organized into two main categories: (i) \textit{\bench{} Generation and Validation Prompts}, which guide the sequential generation and interleaved validation of hierarchical question-answer chains to ensure logical coherence across \lperc{}, \lbridge{}, and \lconn{}, and (ii) \textit{MCTS-Driven Data Generation Pipeline Prompts}, which support the tree search process for exploring diverse reasoning paths, including node generation at each hierarchical level and quality evaluation of candidate question-answer pairs.

\paragraph{List of Prompts}

\begin{enumerate}
  \item \textit{\bench{} Generation and Validation Prompts}
  \begin{enumerate}
    \item \textit{\lconn{} Generation} \dotfill \pageref{fig:prompt_l3analysis}
    \item \textit{\lbridge{} Generation} \dotfill \pageref{fig:prompt_l2generation}
    \item \textit{\lperc{} Generation} \dotfill \pageref{fig:prompt_l1generation}
    \item \textit{\lbridge{} $\rightarrow$ \lconn{} Validation} \dotfill \pageref{fig:prompt_validation_l2_l3}
    \item \textit{\lperc{} $\rightarrow$ \lbridge{} Validation} \dotfill \pageref{fig:prompt_validation_l1_l2}
  \end{enumerate}
  \item \textit{MCTS-Driven Data Generation Pipeline Prompts}
  \begin{enumerate}
    \item \textit{Level-1 Node Generation} \dotfill \pageref{fig:prompt_sft_l1_gen}
    \item \textit{Level-2 Node Generation} \dotfill \pageref{fig:prompt_sft_l2_gen}
    \item \textit{Level-3 Node Generation} \dotfill \pageref{fig:prompt_sft_l3_gen}
    \item \textit{Quality Evaluation} \dotfill \pageref{fig:prompt_sft_eval_hierarchical}
  \end{enumerate}
\end{enumerate}

\clearpage

\begin{figure*}[t]
\small
\centering
\textit{\textbf{$\triangleright$ Implication Understanding (Metaphor)}}\\[8pt]

{%
  \setlength{\fboxsep}{1pt}
  \setlength{\fboxrule}{0pt}
  \fbox{\includegraphics[width=0.35\textwidth]{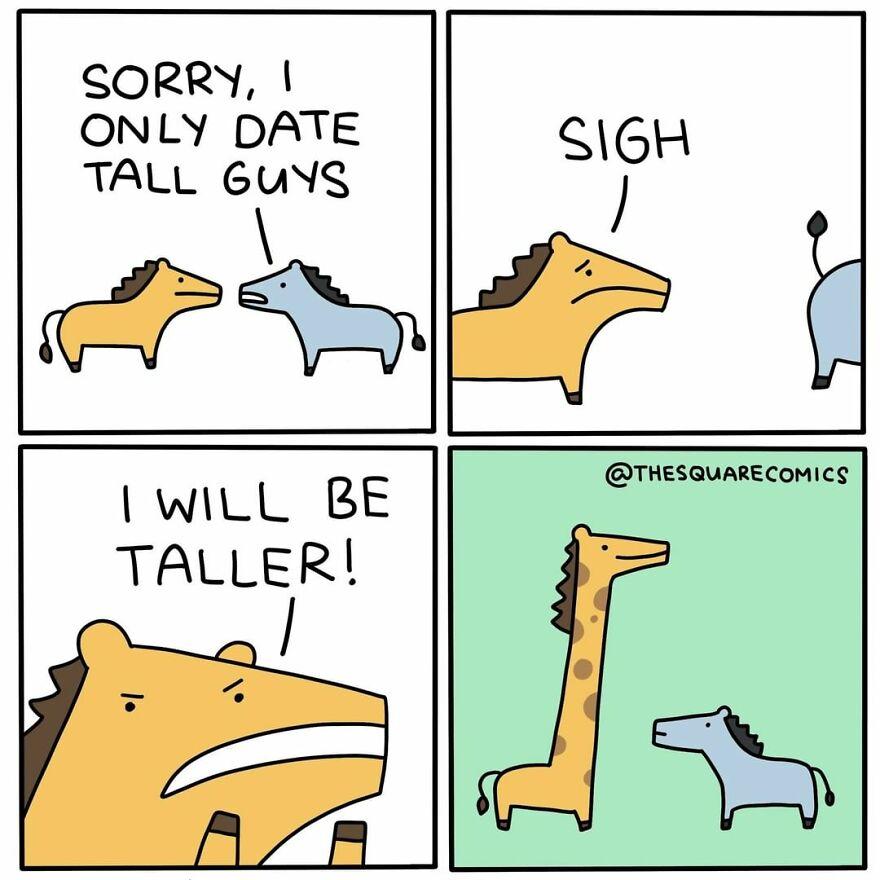}}%
}\\[10pt]

\begin{minipage}{0.95\textwidth}
\raggedright
\footnotesize

\begin{tcolorbox}[enhanced,colback=white,boxrule=0pt,frame style={draw=black, dashed, line width=0.8pt},arc=1pt]
\textbf{Level 1 -- Perception}\\
\textit{Question:} What is the primary method of communication used by the characters in the comic strip?
\begin{enumerate}[label=\textbf{\Alph*}., leftmargin=6mm]
  \item Characters speak dialogue shown in speech bubbles.
  \item They use only non-verbal actions like gestures and facial expressions.
  \item Their communication is shown through thought bubbles above their heads.
  \item A narrator provides descriptions of their thoughts and actions.
\end{enumerate}
\textit{Ground-truth answer:} \textbf{A}.
\end{tcolorbox}

\vspace{2mm}

\begin{tcolorbox}[enhanced,colback=white,boxrule=0pt,frame style={draw=black, dashed, line width=0.8pt},arc=1pt]
\textbf{Level 2 -- Bridge}\\
\textit{Question:} Based on the sequence of events in the comic, what is the most direct reason the yellow horse decides to become taller?
\begin{enumerate}[label=\textbf{\Alph*}., leftmargin=6mm]
  \item To gain a better vantage point for observing its surroundings.
  \item To explore its own potential for extreme physical transformation.
  \item To satisfy the blue horse's stated dating preference for tall individuals.
  \item To develop a more imposing physique for self-defense.
\end{enumerate}
\textit{Ground-truth answer:} \textbf{C}.
\end{tcolorbox}

\vspace{2mm}

\begin{tcolorbox}[enhanced,colback=white,boxrule=0pt,frame style={draw=black, dashed, line width=0.8pt},arc=1pt]
\textbf{Level 3 -- Connotation}\\
\textit{Question:} What could the abrupt transformation of the shorter character into a giraffe in the final panel of the comic strip symbolize in terms of social commentary?
\begin{enumerate}[label=\textbf{\Alph*}., leftmargin=6mm]
  \item It symbolizes the importance of personal growth and improvement.
  \item It represents a critique of the quest for physical perfection and the extremes to which people will go to achieve it.
  \item The transformation illustrates the beauty of embracing one’s unique nature rather than conforming.
  \item The character symbolizes the absurdity of changing oneself to meet others' arbitrary standards.
\end{enumerate}
\textit{Ground-truth answer:} \textbf{D}.
\end{tcolorbox}

\end{minipage}

\caption{A sample from the \textit{Metaphor} aspect of \textit{Implication Understanding}.}
\label{fig:implication_metaphor}
\end{figure*}

\begin{figure*}[t]
\small
\centering
\textit{\textbf{$\triangleright$ Implication Understanding (Contrast)}}\\[8pt]

{%
  \setlength{\fboxsep}{1pt}
  \setlength{\fboxrule}{0pt}
  \fbox{\includegraphics[width=0.35\textwidth]{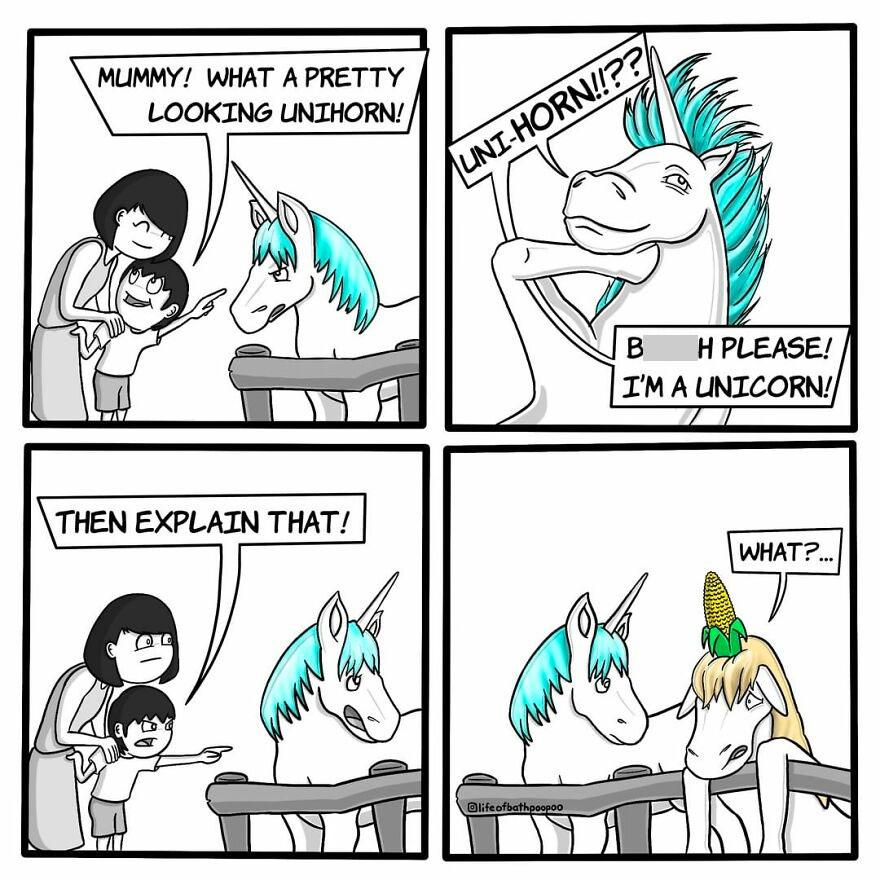}}%
}\\[10pt]

\begin{minipage}{0.95\textwidth}
\raggedright
\footnotesize

\begin{tcolorbox}[
    enhanced,
    colback=white,
    boxrule=0pt,
    frame style={draw=black, dashed, line width=0.8pt},
    arc=1pt,
    left=2pt,right=2pt,top=2pt,bottom=2pt
]
\textbf{Level 1 -- Perception}\\
\textit{Question:} What object is clearly visible on the head of one of the horse-like creatures in the final panel of the comic strip?\\[1mm]
\textit{Options:}
\begin{enumerate}[label=\textbf{\Alph*}., leftmargin=6mm, itemsep=0.3mm, topsep=0.2mm]
  \item A flower.
  \item A small hat.
  \item A unicorn horn.
  \item An ear of corn.
\end{enumerate}
\textit{Ground-truth answer:} \textbf{D}.
\end{tcolorbox}

\vspace{2mm}

\begin{tcolorbox}[
    enhanced,
    colback=white,
    boxrule=0pt,
    frame style={draw=black, dashed, line width=0.8pt},
    arc=1pt,
    left=2pt,right=2pt,top=2pt,bottom=2pt
]
\textbf{Level 2 -- Bridge}\\
\textit{Question:} What is the primary misunderstanding that occurs between the child and the ``unicorn'' character in the initial panels of the comic strip?\\[1mm]
\textit{Options:}
\begin{enumerate}[label=\textbf{\Alph*}., leftmargin=6mm, itemsep=0.3mm, topsep=0.2mm]
  \item The child is referring to a different animal with an actual corn ``horn,'' but the ``unicorn'' character believes the child is misidentifying or insulting its own mythical horn.
  \item The child simply mispronounced the word ``unicorn,'' leading the creature to correct them indignantly.
  \item The ``unicorn'' character mistook the child's comment as a criticism of its horn's appearance, thinking it resembled corn.
  \item The ``unicorn'' character was primarily offended by the child's informal language, rather than the specific word ``unihorn.''
\end{enumerate}
\textit{Ground-truth answer:} \textbf{A}.
\end{tcolorbox}

\vspace{2mm}

\begin{tcolorbox}[
    enhanced,
    colback=white,
    boxrule=0pt,
    frame style={draw=black, dashed, line width=0.8pt},
    arc=1pt,
    left=2pt,right=2pt,top=2pt,bottom=2pt
]
\textbf{Level 3 -- Connotation}\\
\textit{Question:} What is the underlying message conveyed through the humorous twist in the final panel of the comic strip, where an ear of corn is revealed instead of a traditional unicorn horn?\\[1mm]
\textit{Options:}
\begin{enumerate}[label=\textbf{\Alph*}., leftmargin=6mm, itemsep=0.3mm, topsep=0.2mm]
  \item The humor emphasizes our inability to recognize the ordinary when we are obsessed with the extraordinary.
  \item This humorously indicates that our expectations from mythical tales often overlook the charm and surprise found in nature's simplicities.
  \item It mocks the convention of seeking deep, hidden meanings in every aspect of art by providing an unexpectedly literal twist.
  \item It illustrates the importance of deeper investigation and not taking things at face value.
\end{enumerate}
\textit{Ground-truth answer:} \textbf{A}.
\end{tcolorbox}

\end{minipage}

\caption{A sample from the \textit{Contrast} aspect of \textit{Implication Understanding}.}
\label{fig:implication_unicorn_case}
\end{figure*}

\begin{figure*}[t]
\small
\centering
\textit{\textbf{$\triangleright$ Implication Understanding (Exaggeration)}}\\[8pt]

{%
  \setlength{\fboxsep}{1pt}
  \setlength{\fboxrule}{0pt}
  \fbox{\includegraphics[width=0.35\textwidth]{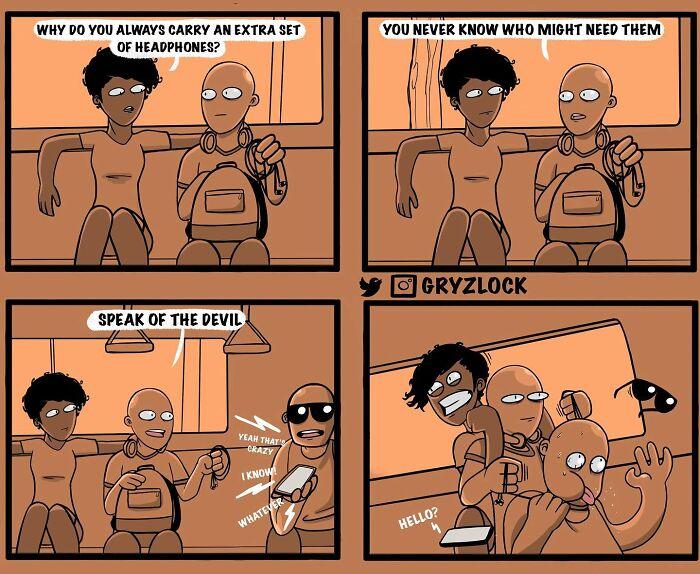}}%
}\\[10pt]

\begin{minipage}{0.95\textwidth}
\raggedright
\footnotesize

\begin{tcolorbox}[enhanced,colback=white,boxrule=0pt,
  frame style={draw=black,dashed,line width=0.8pt},arc=1pt,
  left=2pt,right=2pt,top=2pt,bottom=2pt]
\textbf{Level 1 -- Perception}\\
\textit{Question:} What specific item is the bald character asked about in the first panel, and is visibly holding?\\[1mm]
\textit{Options:}
\begin{enumerate}[label=\textbf{\Alph*}., leftmargin=6mm]
  \item An extra set of headphones.
  \item A mobile phone.
  \item A book.
  \item A pair of sunglasses.
\end{enumerate}
\textit{Ground-truth answer:} \textbf{A}.
\end{tcolorbox}

\vspace{2mm}

\begin{tcolorbox}[enhanced,colback=white,boxrule=0pt,
  frame style={draw=black,dashed,line width=0.8pt},arc=1pt,
  left=2pt,right=2pt,top=2pt,bottom=2pt]
\textbf{Level 2 -- Bridge}\\
\textit{Question:} In the comic, what specific action by the newly arrived character immediately explains why the bald character says ``Speak of the Devil''?\\[1mm]
\textit{Options:}
\begin{enumerate}[label=\textbf{\Alph*}., leftmargin=6mm]
  \item He is holding a smartphone in a public space.
  \item He is talking very loudly on his phone, disturbing the quiet environment.
  \item He is about to ask the bald character for a set of headphones.
  \item He appears to be ignoring the presence of other passengers.
\end{enumerate}
\textit{Ground-truth answer:} \textbf{B}.
\end{tcolorbox}

\vspace{2mm}

\begin{tcolorbox}[enhanced,colback=white,boxrule=0pt,
  frame style={draw=black,dashed,line width=0.8pt},arc=1pt,
  left=2pt,right=2pt,top=2pt,bottom=2pt]
\textbf{Level 3 -- Connotation}\\
\textit{Question:} What does the exaggerated action in the last panel symbolize?\\[1mm]
\textit{Options:}
\begin{enumerate}[label=\textbf{\Alph*}., leftmargin=6mm]
  \item An overblown representation of how technology can suffocate personal freedoms.
  \item The frustrations of public transit riders with loud music.
  \item The extreme discomfort caused by modern technology's invasion into personal space.
  \item A visual hyperbole of the silent plea for etiquette and respect in shared environments.
\end{enumerate}
\textit{Ground-truth answer:} \textbf{D}.
\end{tcolorbox}

\end{minipage}

\caption{A sample from the \textit{Exaggeration} aspect of \textit{Implication Understanding}.}
\label{fig:test108}
\end{figure*}

\begin{figure*}[t]
\small
\centering
\textit{\textbf{$\triangleright$ Implication Understanding (Dislocation)}}\\[8pt]

{%
  \setlength{\fboxsep}{1pt}
  \setlength{\fboxrule}{0pt}
  \fbox{\includegraphics[width=0.35\textwidth]{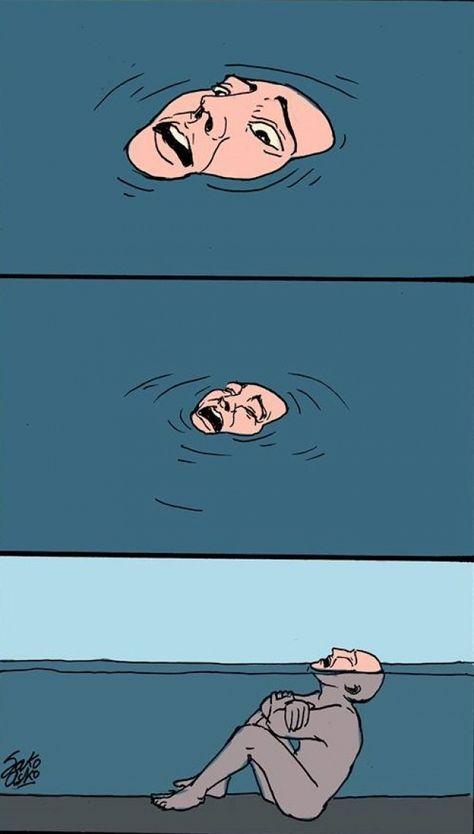}}%
}\\[10pt]

\begin{minipage}{0.95\textwidth}
\raggedright
\footnotesize

\begin{tcolorbox}[enhanced,colback=white,boxrule=0pt,
  frame style={draw=black,dashed,line width=0.8pt},arc=1pt,
  left=2pt,right=2pt,top=2pt,bottom=2pt]
\textbf{Level 1 -- Perception}\\
\textit{Question:} How many distinct panels or frames are present in the image?\\[1mm]
\textit{Options:}
\begin{enumerate}[label=\textbf{\Alph*}., leftmargin=6mm]
  \item One.
  \item Four.
  \item Two.
  \item Three.
\end{enumerate}
\textit{Ground-truth answer:} \textbf{D}.
\end{tcolorbox}

\vspace{2mm}

\begin{tcolorbox}[enhanced,colback=white,boxrule=0pt,
  frame style={draw=black,dashed,line width=0.8pt},arc=1pt,
  left=2pt,right=2pt,top=2pt,bottom=2pt]
\textbf{Level 2 -- Bridge}\\
\textit{Question:} How does the third panel visually recontextualize the seemingly dire situation?\\[1mm]
\textit{Options:}
\begin{enumerate}[label=\textbf{\Alph*}., leftmargin=6mm]
  \item Shows consistent water level and horizon line.
  \item Shows protagonist sitting with crossed arms.
  \item Reveals shallow depth of the water, proving the initial panic was an overreaction.
  \item Shifts to a wider shot including surroundings.
\end{enumerate}
\textit{Ground-truth answer:} \textbf{C}.
\end{tcolorbox}

\vspace{2mm}

\begin{tcolorbox}[enhanced,colback=white,boxrule=0pt,
  frame style={draw=black,dashed,line width=0.8pt},arc=1pt,
  left=2pt,right=2pt,top=2pt,bottom=2pt]
\textbf{Level 3 -- Connotation}\\
\textit{Question:} What hidden meaning might the comic be conveying through the exaggerated expressions followed by the reveal?\\[1mm]
\textit{Options:}
\begin{enumerate}[label=\textbf{\Alph*}., leftmargin=6mm]
  \item It satirizes people's tendency to overreact to non-dangerous situations.
  \item It shows preparation for worst-case scenarios.
  \item It illustrates emotional oscillation between terror and relief.
  \item It mocks sensationalizing everyday events.
\end{enumerate}
\textit{Ground-truth answer:} \textbf{A}.
\end{tcolorbox}

\end{minipage}

\caption{A sample from the \textit{Dislocation} aspect of \textit{Implication Understanding}.}
\label{fig:test1182}
\end{figure*}

\begin{figure*}[t]
\small
\centering
\textit{\textbf{$\triangleright$ Implication Understanding (Symbolism)}}\\[8pt]

{%
\setlength{\fboxsep}{1pt}
\setlength{\fboxrule}{0pt}
\fbox{\includegraphics[width=0.35\textwidth]{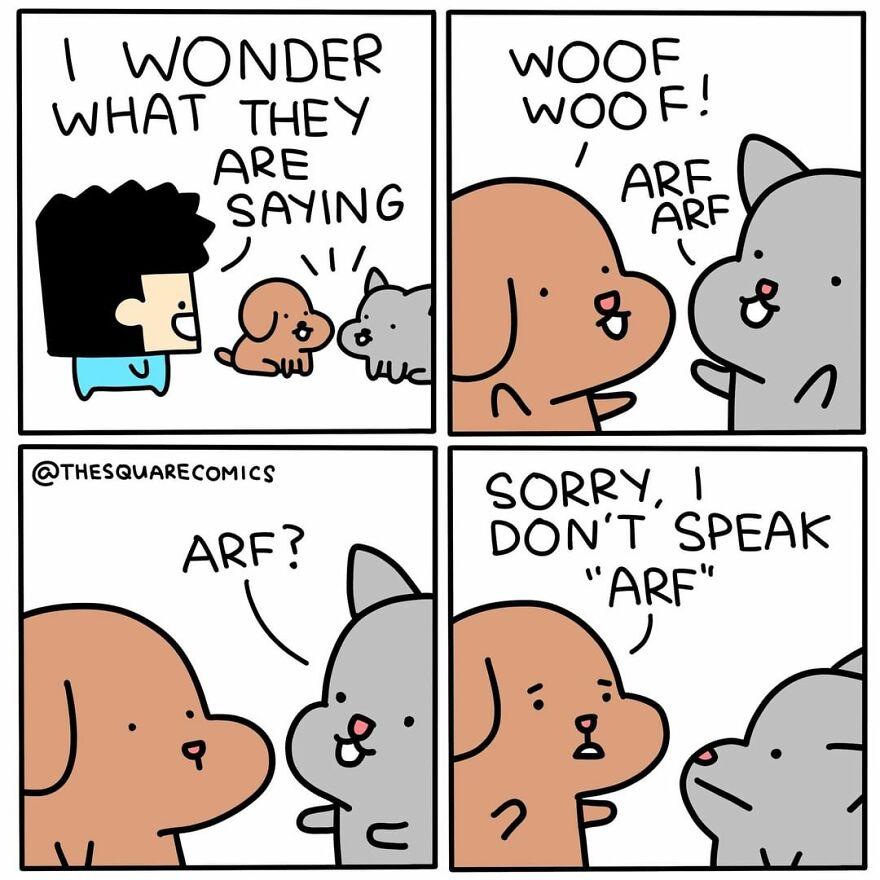}}%
}\\[10pt]

\begin{minipage}{0.95\textwidth}
\raggedright
\footnotesize

\begin{tcolorbox}[enhanced,colback=white,boxrule=0pt,
frame style={draw=black,dashed,line width=0.8pt},arc=1pt,
left=2pt,right=2pt,top=2pt,bottom=2pt]
\textbf{Level 1 -- Perception}\\
\textit{Question:} What animals are depicted interacting in the comic panels?\\[1mm]
\textit{Options:}
\begin{enumerate}[label=\textbf{\Alph*}., leftmargin=6mm]
\item Two dogs.
\item A dog and a cat.
\item A dog and a squirrel.
\item Two cats.
\end{enumerate}
\textit{Ground-truth answer:} \textbf{A}.
\end{tcolorbox}

\vspace{2mm}

\begin{tcolorbox}[enhanced,colback=white,boxrule=0pt,
frame style={draw=black,dashed,line width=0.8pt},arc=1pt,
left=2pt,right=2pt,top=2pt,bottom=2pt]
\textbf{Level 2 -- Bridge}\\
\textit{Question:} What is the source of the humor in the final panel?\\[1mm]
\textit{Options:}
\begin{enumerate}[label=\textbf{\Alph*}., leftmargin=6mm]
\item The brown dog is actually a human in disguise and speaks perfect English.
\item The brown dog claims it cannot ``speak" the specific sound ``ARF") used by the grey dog, treating a simple bark as a foreign language.
\item The grey dog is mute and cannot respond to the brown dog's questions.
\item The human owner misunderstands the barking as a conversation about food.
\end{enumerate}
\textit{Ground-truth answer:} \textbf{B}.
\end{tcolorbox}

\vspace{2mm}

\begin{tcolorbox}[enhanced,colback=white,boxrule=0pt,
frame style={draw=black,dashed,line width=0.8pt},arc=1pt,
left=2pt,right=2pt,top=2pt,bottom=2pt]
\textbf{Level 3 -- Connotation}\\
\textit{Question:} What social phenomenon is implied by the ``language barrier" between the two dogs?\\[1mm]
\textit{Options:}
\begin{enumerate}[label=\textbf{\Alph*}., leftmargin=6mm]
\item The biological inability of different species to communicate.
\item The tendency for people to talk past each other when angry.
\item Even within similar groups, arbitrary linguistic or cultural differences can create barriers to communication.
\item Pets often mimic the behavior and language of their owners.
\end{enumerate}
\textit{Ground-truth answer:} \textbf{C}.
\end{tcolorbox}

\end{minipage}

\caption{A sample from the \textit{Symbolism} aspect of \textit{Implication Understanding}.}
\label{fig:test873}
\end{figure*}

\begin{figure*}[t]
\small
\centering
\textit{\textbf{$\triangleright$ Affective Reasoning (Fear)}}\\[8pt]

{%
  \setlength{\fboxsep}{1pt}
  \setlength{\fboxrule}{0pt}
  \fbox{\includegraphics[width=0.35\textwidth]{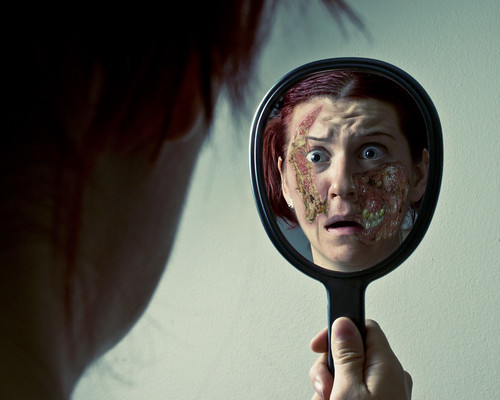}}%
}\\[10pt]

\begin{minipage}{0.95\textwidth}
\raggedright
\footnotesize

\begin{tcolorbox}[
    enhanced,
    colback=white,
    boxrule=0pt,
    frame style={draw=black, dashed, line width=0.8pt},
    arc=1pt,
    left=2pt, right=2pt, top=2pt, bottom=2pt
]
\textbf{Level 1 -- Perception}\\
\textit{Question:} What object is the person in the image holding?\\[1mm]
\textit{Options:}
\begin{enumerate}[label=\textbf{\Alph*}., leftmargin=6mm]
  \item A smartphone.
  \item A hand mirror.
  \item A magnifying glass.
  \item A framed photograph.
\end{enumerate}
\textit{Ground-truth answer:} \textbf{B}.
\end{tcolorbox}

\vspace{2mm}

\begin{tcolorbox}[
    enhanced,
    colback=white,
    boxrule=0pt,
    frame style={draw=black, dashed, line width=0.8pt},
    arc=1pt,
    left=2pt, right=2pt, top=2pt, bottom=2pt
]
\textbf{Level 2 -- Bridge}\\
\textit{Question:} What is the primary emotional dynamic created by the interplay between the woman's facial expression and the visible injuries in the mirror?\\[1mm]
\textit{Options:}
\begin{enumerate}[label=\textbf{\Alph*}., leftmargin=6mm]
  \item The woman's look of surprise indicates she is seeing the injury for the first time, which makes the viewer feel like a witness to a tragic discovery.
  \item The woman's expression of horror validates and amplifies the viewer's reaction to the injury, creating a feedback loop of shared shock and disgust.
  \item The gruesome injuries are the sole source of the viewer's disgust, and the woman's expression simply serves to confirm the reality of the situation.
  \item The woman's horrified expression creates empathy, which conflicts with the feeling of aversion caused by the injury, resulting in a confusing mix of pity and disgust.
\end{enumerate}
\textit{Ground-truth answer:} \textbf{B}.
\end{tcolorbox}

\vspace{2mm}

\begin{tcolorbox}[
    enhanced,
    colback=white,
    boxrule=0pt,
    frame style={draw=black, dashed, line width=0.8pt},
    arc=1pt,
    left=2pt, right=2pt, top=2pt, bottom=2pt
]
\textbf{Level 3 -- Connotation}\\
\textit{Question:} You think the emotional transfer of this picture is perceived as either direct or indirect? Please explain your perspective.\\[1mm]
\textit{Options:}
\begin{enumerate}[label=\textbf{\Alph*}., leftmargin=6mm]
  \item The emotional conveyance is direct, but it primarily communicates feelings of surprise and sadness. The composition suggests the character is seeing the scar for the first time, leading to a moment of shocking self-discovery. The viewer directly experiences this surprise through the close-up perspective. The primary emotional impact is not fear, but a sudden, shared realization of a tragic disfigurement.
  \item The emotional transfer is indirect as it functions on a symbolic level. The magnifying glass represents societal judgment and scrutiny, and the scar symbolizes past trauma. The viewer does not react to the physical image itself but to the abstract concepts of judgment and suffering. This requires an intellectual interpretation of the symbolism before any emotional response is triggered, making the conveyance indirect and contemplative.
  \item The emotional conveyance of this image is direct. It showcases a close-up of the character's face through a hand mirror, particularly highlighting the scar on the face. This emphasis on detail immediately draws the viewer's attention and sympathy towards the character's experiences, while simultaneously evoking fear and aversion towards the horrifying facial details. The character's expression and the use of the hand mirror enhance the visual impact, allowing viewers to quickly grasp the emotions the image intends to convey. This direct visual technique effectively captures the viewer's attention and prompts them to reflect on the character's inner emotions.
  \item The emotional transfer is indirect because it relies on the viewer to construct a narrative. The magnifying glass suggests a story of investigation or scientific study, and the scar is a clue. The emotion is not felt immediately but is deduced after the viewer contemplates the character's potential backstory and the events that led to the injury. This cognitive process of storytelling makes the emotional experience indirect and intellectual.
\end{enumerate}
\textit{Ground-truth answer:} \textbf{C}.
\end{tcolorbox}

\end{minipage}

\caption{A sample from the \textit{Fear} aspect of \textit{Affective Reasoning}.}
\label{fig:test122}
\end{figure*}

\begin{figure*}[t]
\small
\centering
\textit{\textbf{$\triangleright$ Affective Reasoning (Joy)}}\\[8pt]

{%
  \setlength{\fboxsep}{1pt}
  \setlength{\fboxrule}{0pt}
  \fbox{\includegraphics[width=0.35\textwidth]{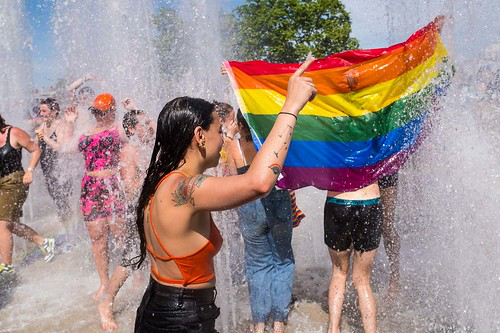}}%
}\\[10pt]

\begin{minipage}{0.95\textwidth}
\raggedright
\footnotesize

\begin{tcolorbox}[enhanced,colback=white,boxrule=0pt,frame style={draw=black, dashed, line width=0.8pt},arc=1pt]
\textbf{Level 1 -- Perception}\\
\textit{Question:} What prominent, colorful object is being held up by the people amidst the splashing water?
\begin{enumerate}[label=\textbf{\Alph*}., leftmargin=6mm]
  \item A large, open umbrella.
  \item A large rainbow flag.
  \item A multi-colored beach towel.
  \item A colorful plastic tarp.
\end{enumerate}
\textit{Ground-truth answer:} \textbf{B}.
\end{tcolorbox}

\vspace{2mm}

\begin{tcolorbox}[enhanced,colback=white,boxrule=0pt,frame style={draw=black, dashed, line width=0.8pt},arc=1pt]
\textbf{Level 2 -- Bridge}\\
\textit{Question:} Why is the atmosphere in the image best characterized as celebratory, rather than simply playful or disorganized?
\begin{enumerate}[label=\textbf{\Alph*}., leftmargin=6mm]
  \item The participants' uninhibited splashing and visible smiles are the primary drivers of the mood, indicating a scene of spontaneous, individual fun.  
  \item The combination of a prominent flag and the occupation of a public fountain suggests a demonstration, where the energy is more disruptive than joyful.
  \item The large rainbow flag provides a shared symbol that gives the energetic activity in the water a sense of collective purpose and joyous expression. 
  \item The energy stems from the refreshing relief the fountain provides on what appears to be a hot day, making the activity a practical response to the environment.
\end{enumerate}
\textit{Ground-truth answer:} \textbf{C}.
\end{tcolorbox}

\vspace{2mm}

\begin{tcolorbox}[enhanced,colback=white,boxrule=0pt,frame style={draw=black, dashed, line width=0.8pt},arc=1pt]
\textbf{Level 3 -- Connotation}\\
\textit{Question:} Which specific elements in the image (such as color, setting, people, etc.) triggered your several main emotional responses? please provide a detailed explanation.
\begin{enumerate}[label=\textbf{\Alph*}., leftmargin=6mm]
  \item My emotional response is driven by the overcast sky and the muted color palette of the people's clothing. The grey, cloudy sky casts a somber light over the scene, suggesting a melancholic or pensive mood. The lack of vibrant colors in the attire, apart from the flag, contributes to a sense of uniformity and blandness. These elements together create a feeling of calmness bordering on sadness, reflecting a subdued and uneventful day. 
  \item The vibrant rainbow flag and the splashing water in the image evoke a strong emotional response. The rainbow-colored flag adds vividness and liveliness to the scene, catching the eye and conveying a celebratory atmosphere through its prominent position in the frame. The splashing water enhances the sense of energy and joy, as people play freely and carefreely in the water. Together, these elements create a positive, inclusive, and celebratory mood, bringing feelings of delight and surprise.  
  \item The image evokes a sense of anger and frustration due to the apparent disorder. The sight of people wading in a public fountain, combined with the prominent display of a flag, suggests a protest or public disruption. This disregard for public property and order triggers a negative response, as it feels chaotic and disrespectful. The splashing water, in this context, is not playful but rather a sign of recklessness, leading to feelings of irritation.  
  \item The primary emotional triggers are the dense crowd and the architectural style of the background buildings. The large number of people packed together evokes a sense of claustrophobia and anxiety, while the imposing, formal architecture creates a feeling of being small and insignificant. This combination leads to a feeling of unease and social discomfort, as the scene appears overwhelming and impersonal.
\end{enumerate}
\textit{Ground-truth answer:} \textbf{B}.
\end{tcolorbox}

\end{minipage}

\caption{A sample from the \textit{Joy} aspect of \textit{Affective Reasoning}.}
\label{fig:test9}
\end{figure*}

\begin{figure*}[t]
\small
\centering
\textit{\textbf{$\triangleright$ Affective Reasoning (Wonder)}}\\[8pt]

{%
  \setlength{\fboxsep}{1pt}
  \setlength{\fboxrule}{0pt}
  \fbox{\includegraphics[width=0.35\textwidth]{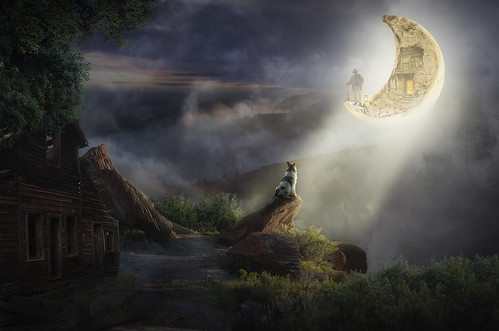}}%
}\\[10pt]

\begin{minipage}{0.95\textwidth}
\raggedright
\footnotesize

\begin{tcolorbox}[enhanced,colback=white,boxrule=0pt,frame style={draw=black, dashed, line width=0.8pt},arc=1pt]
\textbf{Level 1 -- Perception}\\
\textit{Question:} What kind of animal is sitting on the rock in the center of the image, looking up towards the sky?
\begin{enumerate}[label=\textbf{\Alph*}., leftmargin=6mm]
  \item A wolf.
  \item A fox.
  \item A dog.
  \item A cat.
\end{enumerate}
\textit{Ground-truth answer:} \textbf{C}.
\end{tcolorbox}

\vspace{2mm}

\begin{tcolorbox}[enhanced,colback=white,boxrule=0pt,frame style={draw=black, dashed, line width=0.8pt},arc=1pt]
\textbf{Level 2 -- Bridge}\\
\textit{Question:} How does the use of light in the image primarily establish the relationship between the earthly foreground and the fantastical moon?
\begin{enumerate}[label=\textbf{\Alph*}., leftmargin=6mm]
  \item The diffused glow illuminating the mist and mountains establishes a realistic nocturnal setting, creating a dominant atmosphere of quiet serenity and peace.

  \item "The light selectively leaves the abandoned cabin in deep shadow, making its decay the focal point and evoking a primary sense of melancholy and loss.

  \item By casting a direct, focused beam from the moon to the dog, the light creates a narrative link that transforms physical distance into a moment of connection and wonder.

  \item The sharp contrast between the dark landscape and the bright moon serves to isolate the two realms, emphasizing a feeling of insurmountable distance and loneliness.
\end{enumerate}
\textit{Ground-truth answer:} \textbf{C}.
\end{tcolorbox}

\vspace{2mm}

\begin{tcolorbox}[enhanced,colback=white,boxrule=0pt,frame style={draw=black, dashed, line width=0.8pt},arc=1pt]
\textbf{Level 3 -- Connotation}\\
\textit{Question:} Which specific elements (for example, colors, scenes, people, etc.) in the image do you find to evoke the several main emotional reactions from you? please provide a detailed explanation.
\begin{enumerate}[label=\textbf{\Alph*}., leftmargin=6mm]
  \item Several elements in this image evoke strong emotional responses. First, the colors: the dark tones in the scene contrast sharply with the soft moonlight, creating a mysterious and serene atmosphere that catches the eye and sparks a sense of wonder. Next, the setting: the abandoned wooden cabin, rugged mountain rocks, and the distant moon form a dreamlike scene, evoking feelings of both loneliness and longing. Finally, the characters: the dog sitting on the rock and the silhouette of a person on the moon add a narrative quality to the image, as if telling a distant and ancient story. 

  \item The image primarily evokes a sense of fear and anxiety. The dark, ominous tones combined with the dilapidated, abandoned cabin suggest a scene of horror or danger. The rugged, sharp rocks and the isolated setting create a feeling of vulnerability and entrapment. The dog appears tense, as if sensing a threat, and the silhouette on the moon is a menacing, watchful figure. This combination of elements fosters a deep sense of unease and foreboding, making the viewer feel tense and scared.

  \item This image communicates overwhelming sadness and melancholy. The abandoned cabin is a clear symbol of loss and forgotten times, evoking a deep sense of grief. The barren, rocky landscape emphasizes the feeling of utter loneliness and despair. The single dog, sitting alone, appears to be waiting for a companion who will never return, adding a layer of poignant sorrow. The moonlight casts a cold, mournful glow, amplifying the feeling of desolation and making the entire scene feel tragic.

  \item The image evokes a sense of vibrant energy and excitement. The sharp contrast between the light and dark areas creates a dynamic, high-energy feeling, like a flash of lightning. The rugged mountains suggest adventure and challenge, sparking a desire for action. The cabin, while old, looks like a basecamp for an exciting exploration. The dog is poised as if about to spring into action, and the figure on the moon appears to be a triumphant explorer. The entire scene feels like the beginning of a thrilling journey, filling the viewer with anticipation and exhilaration.
\end{enumerate}
\textit{Ground-truth answer:} \textbf{A}.
\end{tcolorbox}

\end{minipage}

\caption{A sample from the \textit{Wonder} aspect of \textit{Affective Reasoning}.}
\label{fig:test270}
\end{figure*}

\begin{figure*}[t]
\small
\centering
\textit{\textbf{$\triangleright$ Affective Reasoning (Anger)}}\\[8pt]

{%
  \setlength{\fboxsep}{1pt}
  \setlength{\fboxrule}{0pt}
  \fbox{\includegraphics[width=0.35\textwidth]{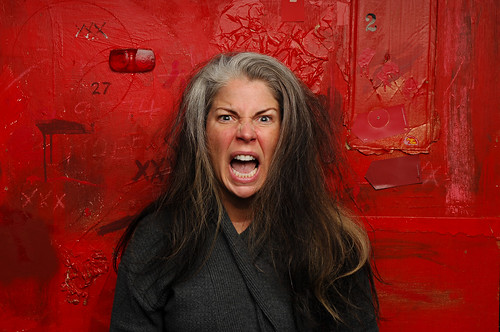}}%
}\\[10pt]

\begin{minipage}{0.95\textwidth}
\raggedright
\footnotesize

\begin{tcolorbox}[
  enhanced, colback=white, boxrule=0pt,
  frame style={draw=black, dashed, line width=0.8pt},
  arc=1pt]
\textbf{Level 1 -- Perception}\\
\textit{Question:} What primary emotion is the woman's facial expression conveying?
\begin{enumerate}[label=\textbf{\Alph*}., leftmargin=6mm]
  \item Sadness.
  \item Anger.
  \item Surprise.
  \item Joy.
\end{enumerate}
\textit{Ground-truth answer:} \textbf{B}.
\end{tcolorbox}

\vspace{2mm}

\begin{tcolorbox}[
  enhanced, colback=white, boxrule=0pt,
  frame style={draw=black, dashed, line width=0.8pt},
  arc=1pt]
\textbf{Level 2 -- Bridge}\\
\textit{Question:} How does the background interact with the woman’s expression?
\begin{enumerate}[label=\textbf{\Alph*}., leftmargin=6mm]
  \item The intense red of the background amplifies the raw anger in her expression, creating a more powerful and overwhelming feeling of rage.

  \item The red background is the primary source of aggression, causing the woman's expression to be interpreted as a reaction of fear or feeling cornered.

  \item The flat, theatrical quality of the red background contrasts with the realistic expression, creating an emotional ambiguity between genuine rage and a staged performance.

  \item The chaotic details in the red background, like graffiti and decay, suggest a specific narrative cause for her anger, making it feel targeted at her surroundings.
\end{enumerate}
\textit{Ground-truth answer:} \textbf{A}.
\end{tcolorbox}

\vspace{2mm}

\begin{tcolorbox}[
  enhanced, colback=white, boxrule=0pt,
  frame style={draw=black, dashed, line width=0.8pt},
  arc=1pt]
\textbf{Level 3 -- Connotation}\\
\textit{Question:} What elements create the emotional responses you feel?
\begin{enumerate}[label=\textbf{\Alph*}., leftmargin=6mm]
  \item The image primarily conveys a profound sense of sadness and despair. The character's wide-open mouth is not a scream of anger but a cry of anguish and loss. The stark red background isolates her, symbolizing her inner pain and emotional turmoil, trapping her in her grief. Her wide eyes are filled with hopelessness, not rage. The entire composition communicates a deep, personal tragedy, making the viewer feel empathy and a shared sense of sorrow for her suffering.

  \item The scene appears to be one of theatrical performance, which inspires a sense of awe and surprise. The red is reminiscent of a stage curtain, suggesting a dramatic reveal, and the character's expression is one of pure astonishment, as if witnessing something incredible for the first time. This creates a feeling of wonder and anticipation in the viewer, rather than anger or fear. The scene feels exciting and spectacular, drawing the viewer in with a sense of dramatic tension and curiosity about what she is seeing.

  \item The primary emotional trigger is the lack of a detailed environment, focusing solely on the character against a plain background. This minimalism creates a sense of loneliness and isolation. The character's expression seems to be one of confusion and vulnerability, as if she is lost or abandoned. The red color does not feel aggressive but rather empty and vast, amplifying her solitude. The main emotions evoked are therefore pity and a gentle melancholy for the character's apparent plight.

  \item The intense red background and the expression of the character in the image evoke the strongest emotional reaction in me. Red is often associated with passion and power, and here it may symbolize anger or intense emotion. Combined with the character's expression, particularly her wide-open mouth and wide eyes, it conveys rage, fury, or extreme dissatisfaction. This sense of anger is also transmitted to the viewer, becoming the primary emotion evoked. The combination of this expression with the red background enhances the overall tension and unease, creating a sense of oppression that makes the viewer feel fear. Additionally, the intense visual content and color scheme can cause slight discomfort, thereby eliciting a sense of aversion.
\end{enumerate}
\textit{Ground-truth answer:} \textbf{D}.
\end{tcolorbox}

\end{minipage}

\caption{A sample from the \textit{Anger} aspect of \textit{Affective Reasoning}.}
\label{fig:test257}
\end{figure*}

\begin{figure*}[t]
\small
\centering
\textit{\textbf{$\triangleright$ Affective Reasoning (Affection)}}\\[8pt]

{%
  \setlength{\fboxsep}{1pt}
  \setlength{\fboxrule}{0pt}
  \fbox{\includegraphics[width=0.35\textwidth]{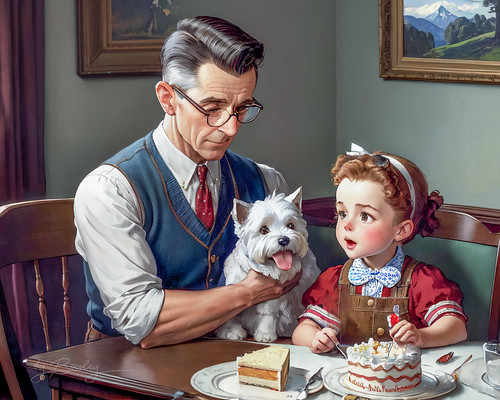}}%
}\\[10pt]

\begin{minipage}{0.95\textwidth}
\raggedright
\footnotesize

\begin{tcolorbox}[enhanced,colback=white,boxrule=0pt,frame style={draw=black, dashed, line width=0.8pt},arc=1pt]
\textbf{Level 1 -- Perception}\\
\textit{Question:} What three main subjects are gathered at the table?
\begin{enumerate}[label=\textbf{\Alph*}., leftmargin=6mm]
  \item A man and a young girl.  
  \item A man, a young girl, and a dog.  
  \item A man, a young girl, and a cat.  
  \item A woman, a young boy, and a dog.
\end{enumerate}
\textit{Ground-truth answer:} \textbf{B}.
\end{tcolorbox}

\vspace{2mm}

\begin{tcolorbox}[enhanced,colback=white,boxrule=0pt,frame style={draw=black, dashed, line width=0.8pt},arc=1pt]
\textbf{Level 2 -- Bridge}\\
\textit{Question:} How do their actions and expressions establish the emotional tone?
\begin{enumerate}[label=\textbf{\Alph*}., leftmargin=6mm]
  \item The father's expression of gentle nostalgia as he looks down suggests a bittersweet moment, while the daughter's wide-eyed look shows her concern for his feelings.

  \item The father's gentle gaze towards the girl, combined with her look of happy surprise and the presence of a birthday cake, creates a shared moment of familial joy.
  
  \item The scene's happiness stems primarily from the dog, which is being presented as a gift, causing the daughter's excitement and the father's look of satisfaction.

  \item The daughter's expression of eager anticipation is directed at the cake, while the father's observant look suggests he is waiting for her reaction, creating a sense of suspense.
\end{enumerate}
\textit{Ground-truth answer:} \textbf{B}.
\end{tcolorbox}

\vspace{2mm}

\begin{tcolorbox}[enhanced,colback=white,boxrule=0pt,frame style={draw=black, dashed, line width=0.8pt},arc=1pt]
\textbf{Level 3 -- Connotation}\\
\textit{Question:} Describe the emotional content (valence, arousal, dominance).
\begin{enumerate}[label=\textbf{\Alph*}., leftmargin=6mm]
  \item The picture depicts a father and daughter in a moment of disagreement. The father, with a stern expression, holds the dog back, which appears agitated. The daughter looks away from the cake, her face showing disappointment and sadness. The cartoon style contrasts with the underlying tension. The main emotions are sadness, frustration, and tension. This scene has a low valence, a high level of arousal due to the conflict, and a low level of dominance as the characters feel constrained by the negative situation.

  \item The picture depicts a scene where the father and the family's little dog are celebrating the daughter's birthday together. The father, wearing glasses, looks at his daughter with a kind expression, holding the little dog in his hands. The dog, with its tongue hanging out, gazes at the daughter and the cake, while the daughter looks back at her father and the dog, creating a joyful and heartwarming family atmosphere that brings happiness and delight. On the table, there are two pieces of cake, which are visually appealing, delicious, and surprisingly delightful. The entire scene has been rendered in a cartoonish style, maintaining a certain level of neutrality. The levels of joy and dominance are both high, evoking feelings of happiness and surprise. The arousal level is moderate, reflecting a warm and positive emotional tone.

  \item The image portrays a family preparing for a pet competition. The father is holding their small dog, showcasing it to his daughter who acts as a judge. The dog seems eager, while the daughter's expression is one of serious concentration. The cakes on the table are rewards. The primary emotions are anticipation, concentration, and a sense of pressure. This results in a moderate valence, high arousal due to the competitive tension, and high dominance reflecting the characters' focus and control.

  \item This is a scene of a family saying goodbye. The father is about to leave and is holding the family dog for his daughter to pet one last time. The daughter looks up at him with a sad expression. The cake on the table is a farewell treat. The emotions conveyed are sadness, love, and longing. The cartoon rendering softens the sad theme but doesn't erase it. The valence is low, arousal is moderate due to the poignant emotions, and dominance is low, signifying a sense of loss of control over events.
\end{enumerate}
\textit{Ground-truth answer:} \textbf{B}.
\end{tcolorbox}

\end{minipage}

\caption{A sample from the \textit{Affection} aspect of \textit{Affective Reasoning}.}
\label{fig:test117}
\end{figure*}

\begin{figure*}[t]
\small
\centering
\textit{\textbf{$\triangleright$ Affective Reasoning (Sadness)}}\\[8pt]

{%
  \setlength{\fboxsep}{1pt}
  \setlength{\fboxrule}{0pt}
  \fbox{\includegraphics[width=0.35\textwidth]{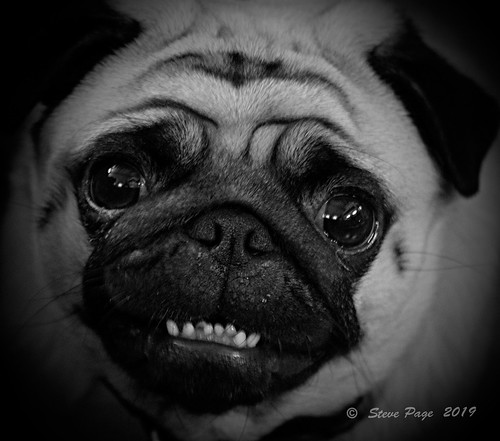}}%
}\\[10pt]

\begin{minipage}{0.95\textwidth}
\raggedright
\footnotesize

\begin{tcolorbox}[
  enhanced, colback=white, boxrule=0pt,
  frame style={draw=black, dashed, line width=0.8pt},
  arc=1pt, left=2pt,right=2pt,top=2pt,bottom=2pt]
\textbf{Level 1 -- Perception}\\
\textit{Question:} What type of animal is the main subject of this close-up photograph?
\begin{enumerate}[label=\textbf{\Alph*}., leftmargin=6mm]
  \item A monkey.
  \item A cat.
  \item A dog.
  \item A bear cub.
\end{enumerate}
\textit{Ground-truth answer:} \textbf{C}.
\end{tcolorbox}

\vspace{2mm}

\begin{tcolorbox}[
  enhanced, colback=white, boxrule=0pt,
  frame style={draw=black, dashed, line width=0.8pt},
  arc=1pt, left=2pt,right=2pt,top=2pt,bottom=2pt]
\textbf{Level 2 -- Bridge}\\
\textit{Question:} How do the different parts of the pug's facial expression interact to create a sense of emotional ambiguity?
\begin{enumerate}[label=\textbf{\Alph*}., leftmargin=6mm]
  \item The primary source of ambiguity lies solely within the pug's eyes, which appear both large and curious due to light reflection while also having a sad, downward-turned shape.

  \item The expression is a result of the pug's physical structure; the bared teeth are a common trait of the breed's underbite and not an emotional sign, while the wide eyes indicate a state of high alert or fear.

  \item The deep wrinkles on the pug's forehead, suggesting worry, conflict with the dramatic, high-contrast lighting, which gives the image an aggressive and menacing quality.

  \item The pug's wide, seemingly sad eyes contrast with its bared teeth, which could be interpreted as either aggression or a playful smile, creating an uncertain emotional signal.
\end{enumerate}
\textit{Ground-truth answer:} \textbf{D}.
\end{tcolorbox}

\vspace{2mm}

\begin{tcolorbox}[
  enhanced, colback=white, boxrule=0pt,
  frame style={draw=black, dashed, line width=0.8pt},
  arc=1pt, left=2pt,right=2pt,top=2pt,bottom=2pt]
\textbf{Level 3 -- Connotation}\\
\textit{Question:} When you look at this image, do you feel any conflicting emotions or uncertainty?  
\begin{enumerate}[label=\textbf{\Alph*}., leftmargin=6mm]
  \item Upon viewing the image, there is no sense of emotional conflict. The pug's expression is one of clear and unambiguous aggression and anger. The bared teeth are a sign of a snarl, and the tension in its facial muscles indicates hostility. The stark, high-contrast black and white photography amplifies this feeling of menace. The viewer is meant to feel intimidated and fearful, as the dog is asserting its dominance in a threatening manner, leaving no room for emotional ambiguity.

  \item When viewing this image, one might experience an emotional conflict. The expression of the pug in the picture appears somewhat melancholic yet with a hint of playfulness. Its teeth are slightly exposed, as if it's smiling, but there's a touch of seriousness in its eyes. This mixed expression makes it difficult to determine the dog's emotional state, creating a sense of uncertainty that manifests in the viewer primarily holding a neutral emotion. At the same time, the black and white tones add a sense of oppression to the scene, potentially evoking the viewer's curiosity and sympathy towards the dog's emotional state, leading to feelings of sadness and fear.

  \item This image presents a straightforward and non-conflicting emotional state of pure curiosity. The pug's head is tilted, and its eyes are focused, suggesting it is intently observing something just out of frame. The slightly open mouth and exposed teeth are characteristic of a dog's relaxed, inquisitive posture. The black and white tones create a dramatic effect that highlights the intensity of the dog's focus. The viewer primarily feels a sense of engagement and wonder, not emotional conflict or sadness.

  \item There is no emotional uncertainty in this image; it is a clear depiction of physical discomfort and pain. The pug's squinted eyes and bared teeth are not related to any complex emotion but are physical reactions to an unpleasant sensation, such as an injury or illness. The black and white aesthetic lends a somber, clinical quality to the scene, emphasizing the animal's suffering. The viewer's response is one of sympathy and concern, driven by the unambiguous signs of physical distress, not emotional conflict.
\end{enumerate}
\textit{Ground-truth answer:} \textbf{B}.
\end{tcolorbox}

\end{minipage}

\caption{A sample from the \textit{Sadness } aspect of \textit{Affective Reasoning}.}
\label{fig:test30}
\end{figure*}

\begin{figure*}[t]
\small
\centering
\textit{\textbf{$\triangleright$ Aesthetic Appreciation (Graphic)}}\\[8pt]

{%
  \setlength{\fboxsep}{1pt}
  \setlength{\fboxrule}{0pt}
  \fbox{\includegraphics[width=0.35\textwidth]{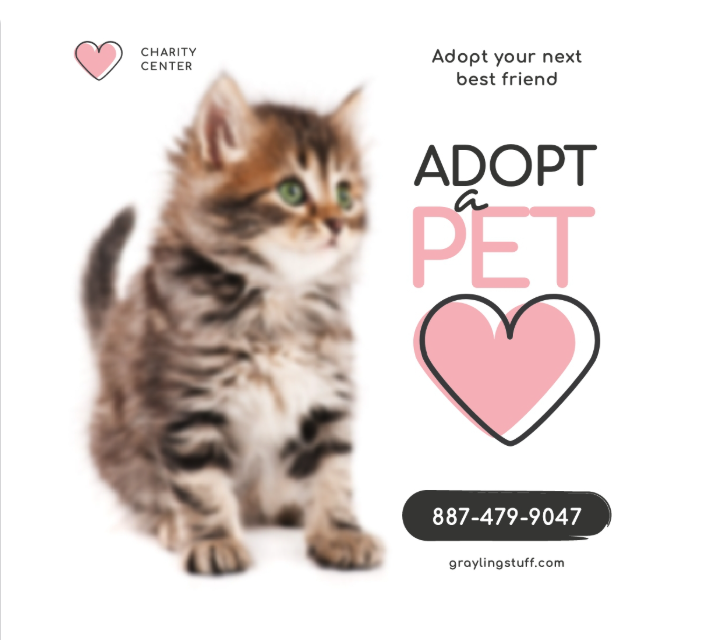}}%
}\\[10pt]

\begin{minipage}{0.95\textwidth}
\raggedright
\footnotesize

\begin{tcolorbox}[
    enhanced, colback=white, boxrule=0pt,
    frame style={draw=black, dashed, line width=0.8pt},
    arc=1pt, left=2pt,right=2pt,top=2pt,bottom=2pt
]
\textbf{Level 1 -- Perception}\\
\textit{Question:} What type of animal is the main subject of this advertisement?

\begin{enumerate}[label=\textbf{\Alph*}., leftmargin=6mm]
\item A rabbit
\item A hamster
\item A puppy
\item A kitten
\end{enumerate}

\textit{Ground-truth answer:} \textbf{D}.
\end{tcolorbox}

\vspace{2mm}

\begin{tcolorbox}[
    enhanced, colback=white, boxrule=0pt,
    frame style={draw=black, dashed, line width=0.8pt},
    arc=1pt, left=2pt,right=2pt,top=2pt,bottom=2pt
]
\textbf{Level 2 -- Bridge}\\
\textit{Question:} By comparing the kitten image to the surrounding text and graphics, what is the primary visual tension or conflict within the ad's composition?

\begin{enumerate}[label=\textbf{\Alph*}., leftmargin=6mm]
\item The blur on the kitten is a deliberate technique to create depth of field, pushing it into the background to make the “ADOPT a PET” text the main focus.
\item There is a conflict in balance; the visually heavy kitten on the left competes for attention with the text block on the right, dividing the viewer's focus.
\item There is a conflict in style; the realistic photograph of the kitten clashes with the flat, illustrative style of the heart graphics and typography.
\item There is a conflict in sharpness; the kitten, the emotional focus, is blurry and indistinct, while the text and logos are crisp and clear.
\end{enumerate}

\textit{Ground-truth answer:} \textbf{D}.
\end{tcolorbox}

\vspace{2mm}

\begin{tcolorbox}[
    enhanced, colback=white, boxrule=0pt,
    frame style={draw=black, dashed, line width=0.8pt},
    arc=1pt, left=2pt,right=2pt,top=2pt,bottom=2pt
]
\textbf{Level 3 -- Connotation}\\
\textit{Question:} Does this design have any issues that affect its effectiveness? If so, what are the main impacts?

\begin{enumerate}[label=\textbf{\Alph*}., leftmargin=6mm]
\item Yes, the lack of sharpness in the main subject significantly weakens the design's effectiveness. Even though the text is legible, this visual flaw undermines the emotional connection with the audience and damages the organization's perceived professionalism.
\item Yes, the inconsistent and overly playful typography creates a sense of disorganization that is difficult to read quickly. This confusing hierarchy detracts from the message's urgency and makes the brand appear less credible.
\item No, the design does not have significant issues; in fact, the soft focus on the subject creates an artistic, dreamy effect that captures attention. This stylistic choice encourages the viewer to focus on the main “Adopt a Pet” message, enhancing its communicative power.
\item The design has a minor issue with image clarity, but its impact is minimal. The strong, legible headline and clear contact information effectively compensate for this, ensuring the core message of adoption is still successfully communicated.
\end{enumerate}

\textit{Ground-truth answer:} \textbf{A}.
\end{tcolorbox}

\end{minipage}

\caption{A sample from the \textit{Graphic} aspect of \textit{Aesthetic Appreciation}.}
\label{fig:graphic240}
\end{figure*}

\begin{figure*}[t]
\small
\centering
\textit{\textbf{$\triangleright$ Aesthetic Appreciation (Color)}}\\[8pt]

{%
  \setlength{\fboxsep}{1pt}
  \setlength{\fboxrule}{0pt}
  \fbox{\includegraphics[width=0.35\textwidth]{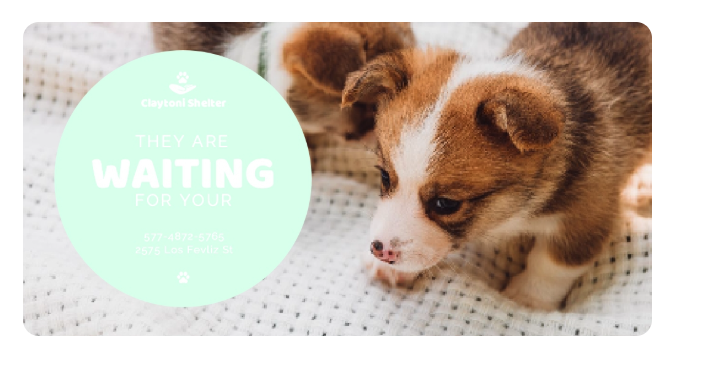}}%
}\\[10pt]

\begin{minipage}{0.95\textwidth}
\raggedright
\footnotesize

\begin{tcolorbox}[enhanced,colback=white,boxrule=0pt,
frame style={draw=black, dashed, line width=0.8pt},arc=1pt]
\textbf{Level 1 -- Perception}\\
\textit{Question:} What is the color of the large circular area on the left side of the image where the text is located?

\begin{enumerate}[label=\textbf{\Alph*}., leftmargin=6mm]
\item Light Green
\item Light Yellow
\item Light Blue
\item White
\end{enumerate}

\textit{Ground-truth answer:} \textbf{A}.
\end{tcolorbox}

\vspace{2mm}

\begin{tcolorbox}[enhanced,colback=white,boxrule=0pt,
frame style={draw=black, dashed, line width=0.8pt},arc=1pt]
\textbf{Level 2 --  Bridge}\\
\textit{Question:} Why might a potential adopter have difficulty reading the contact information for the “Claytoni Shelter” in this advertisement?

\begin{enumerate}[label=\textbf{\Alph*}., leftmargin=6mm]
\item The font size for the contact information is too small compared to the headline, making it the primary reason it is difficult to read.
\item The white text of the address and phone number has very low color contrast against the light green background, making it nearly illegible.
\item The circular text box is placed over a visually complex part of the background image, which interferes with the text.
\item The large word “WAITING” is so visually dominant that it draws the eye away from the contact details, making them hard to find.
\end{enumerate}

\textit{Ground-truth answer:} \textbf{B}.
\end{tcolorbox}

\vspace{2mm}

\begin{tcolorbox}[enhanced,colback=white,boxrule=0pt,
frame style={draw=black, dashed, line width=0.8pt},arc=1pt]
\textbf{Level 3 -- Connotation}\\
\textit{Question:} Does this design have any issues that affect its effectiveness? If so, what are the main impacts?

\begin{enumerate}[label=\textbf{\Alph*}., leftmargin=6mm]
\item No, the design is highly effective and communicates a gentle, caring brand identity. The soft pastel color palette creates a calming, approachable atmosphere that enhances the emotional appeal of the subject matter and strengthens the brand's positioning.
\item While the text is somewhat faint, the overall minimalist aesthetic is clean, modern, and visually pleasing. The subtle color treatment is a reasonable trade-off that supports a non-aggressive tone, resulting in a minor, acceptable impact on readability.
\item Yes, the design's color choices create significant legibility problems that obscure vital information. Even if the palette intends to be soft and gentle, this choice severely undermines communication effectiveness and damages the organization's credibility by appearing unprofessional.
\item Yes, the design's typography feels dated and generic, which fails to capture attention effectively. This weak font choice makes the brand seem uninspired and reduces the overall persuasive power of the message, even with a strong central image.
\end{enumerate}

\textit{Ground-truth answer:} \textbf{C}.
\end{tcolorbox}

\end{minipage}

\caption{A sample from the \textit{Color} aspect of \textit{Aesthetic Appreciation}.}
\label{fig:color164}
\end{figure*}

\begin{figure*}[t]
\small
\centering
\textit{\textbf{$\triangleright$ Aesthetic Appreciation (Font)}}\\[8pt]

{%
  \setlength{\fboxsep}{1pt}
  \setlength{\fboxrule}{0pt}
  \fbox{\includegraphics[width=0.35\textwidth]{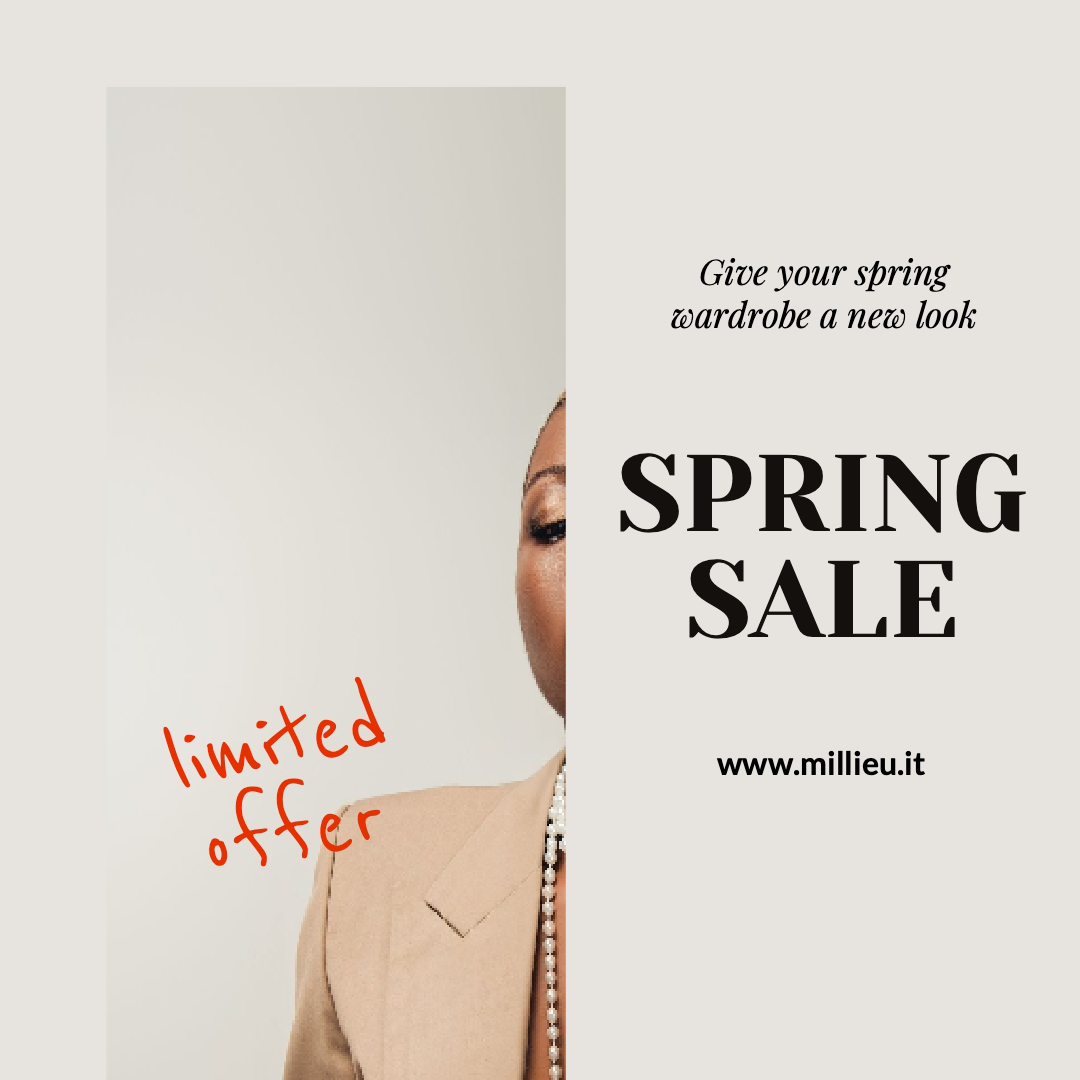}}%
}\\[10pt]

\begin{minipage}{0.95\textwidth}
\raggedright
\footnotesize

\begin{tcolorbox}[enhanced,colback=white,boxrule=0pt,
frame style={draw=black, dashed, line width=0.8pt},arc=1pt]
\textbf{Level 1 -- Perception}\\
\textit{Question:} What color is the handwritten text that reads “limited offer”?

\begin{enumerate}[label=\textbf{\Alph*}., leftmargin=6mm]
\item Beige
\item Black
\item Red
\item White
\end{enumerate}

\textit{Ground-truth answer:} \textbf{C}.
\end{tcolorbox}

\vspace{2mm}

\begin{tcolorbox}[enhanced,colback=white,boxrule=0pt,
frame style={draw=black, dashed, line width=0.8pt},arc=1pt]
\textbf{Level 2 -- Bridge}\\
\textit{Question:} How do the visual characteristics of the “SPRING SALE” text and the “limited offer” text create different impressions for the viewer?

\begin{enumerate}[label=\textbf{\Alph*}., leftmargin=6mm]
\item The “SPRING SALE” text is bold to show importance, while the “limited offer” text is red to serve as a warning to the customer.
\item The “SPRING SALE” text establishes a modern, minimalist theme, while the “limited offer” text adds a personal, artistic touch.
\item Both text elements are designed primarily to grab attention, using different fonts to distinguish the main headline from the secondary condition.
\item The “SPRING SALE” text uses an elegant, formal font suggesting high quality, while the “limited offer” text uses a casual, handwritten font suggesting informal urgency.
\end{enumerate}

\textit{Ground-truth answer:} \textbf{D}.
\end{tcolorbox}

\vspace{2mm}

\begin{tcolorbox}[enhanced,colback=white,boxrule=0pt,
frame style={draw=black, dashed, line width=0.8pt},arc=1pt]
\textbf{Level 3 -- Connotation}\\
\textit{Question:} Does this design present any significant issues affecting its overall effectiveness? If so, what are the primary high-level impacts on communication and brand perception?

\begin{enumerate}[label=\textbf{\Alph*}., leftmargin=6mm]
\item The design has a minor issue with the handwritten font feeling slightly out of place, but its impact is minimal. The clarity of the main “Spring Sale” headline and the high-quality photography are strong enough to maintain the brand's premium positioning and ensure the message is communicated effectively.
\item Yes, the design suffers from an incongruous text element that clashes with the established sophisticated aesthetic. Even though the urgent message is legible, this stylistic mismatch undermines the brand's perceived professionalism and creates a visual dissonance that can cheapen the overall impression.
\item No, the design is highly effective because the mix of formal and informal typography creates a dynamic contrast that captures attention. This stylistic choice makes the promotion feel more accessible and urgent, ultimately boosting engagement without harming the core brand identity.
\item Yes, the placement of the text elements creates a cluttered and unbalanced composition that competes with the main image. This poor spatial organization weakens the visual hierarchy, making it harder for viewers to process the information efficiently and grasp the key message at a glance.
\end{enumerate}

\textit{Ground-truth answer:} \textbf{B}.
\end{tcolorbox}

\end{minipage}

\caption{A sample from the \textit{Font} aspect of \textit{Aesthetic Appreciation}.}
\label{fig:font27}
\end{figure*}

\begin{figure*}[t]
\small
\centering
\textit{\textbf{$\triangleright$ Aesthetic Appreciation (Composition)}}\\[8pt]

{%
  \setlength{\fboxsep}{1pt}
  \setlength{\fboxrule}{0pt}
  \fbox{\includegraphics[width=0.35\textwidth]{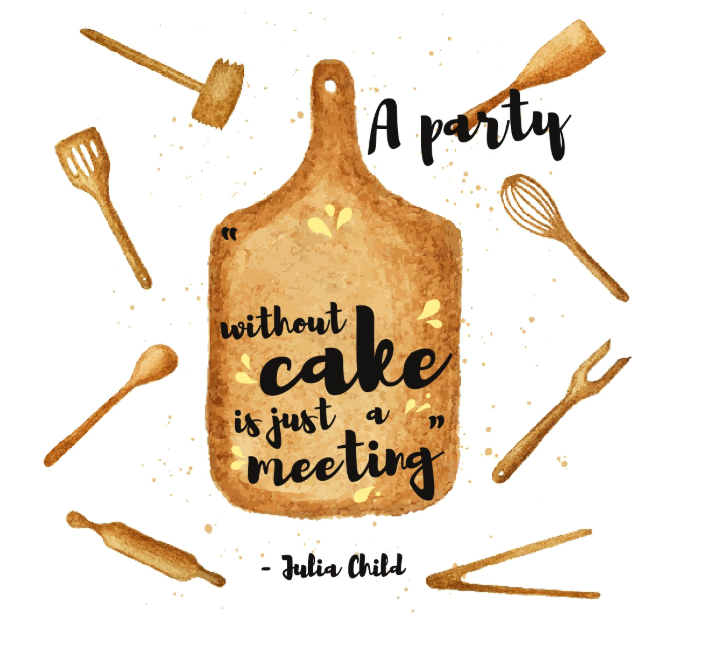}}%
}\\[10pt]

\begin{minipage}{0.95\textwidth}
\raggedright
\footnotesize

\begin{tcolorbox}[enhanced,colback=white,boxrule=0pt,
frame style={draw=black, dashed, line width=0.8pt},arc=1pt]
\textbf{Level 1 -- Perception}\\
\textit{Question:} What is the primary object that serves as the background for most of the text in the image?

\begin{enumerate}[label=\textbf{\Alph*}., leftmargin=6mm]
\item A cutting board
\item A piece of parchment paper
\item A baker's peel
\item A serving platter
\end{enumerate}

\textit{Ground-truth answer:} \textbf{A}.
\end{tcolorbox}

\vspace{2mm}

\begin{tcolorbox}[enhanced,colback=white,boxrule=0pt,
frame style={draw=black, dashed, line width=0.8pt},arc=1pt]
\textbf{Level 2 -- Bridge}\\
\textit{Question:} How is the quote “A party without cake is just a meeting” spatially organized in relation to the cutting board illustration?

\begin{enumerate}[label=\textbf{\Alph*}., leftmargin=6mm]
\item The quote begins inside the top of the cutting board and flows downwards, with only the author's name appearing outside the board.
\item The entire quote is contained within the boundaries of the cutting board, with the most important words enlarged for emphasis.
\item The main subjects, “party” and “cake”, are positioned outside the board to draw attention, while the rest of the phrase is inside.
\item The first two words, “A party”, are positioned outside the cutting board, while the remainder of the quote is located inside it.
\end{enumerate}

\textit{Ground-truth answer:} \textbf{D}.
\end{tcolorbox}

\vspace{2mm}

\begin{tcolorbox}[enhanced,colback=white,boxrule=0pt,
frame style={draw=black, dashed, line width=0.8pt},arc=1pt]
\textbf{Level 3 -- Connotation}\\
\textit{Question:} Does this design exhibit any issues that compromise its overall effectiveness? If so, what are the primary consequences for the viewer's experience and the design's perceived quality?

\begin{enumerate}[label=\textbf{\Alph*}., leftmargin=6mm]
\item No, the design does not have significant issues; in fact, its unconventional typography is effective at capturing attention. This dynamic arrangement adds visual interest and energy, reinforcing the celebratory theme of the quote.
\item While the separation of the first two words is slightly unconventional, it creates a unique visual entry point into the quote. The overall charming, handcrafted aesthetic is strong enough to compensate for this minor quirk, ensuring the design remains effective and endearing.
\item Yes, the muted, monochromatic color palette makes the design feel dated and unexciting. This lack of vibrant color fails to convey the joyful nature of a “party”, thereby weakening the message's emotional impact and reducing its overall memorability.
\item Yes, the design suffers from disjointed text placement that fragments the central message. Even if the playful font is appealing, this structural flaw disrupts reading flow and undermines the design's sense of polish and professionalism.
\end{enumerate}

\textit{Ground-truth answer:} \textbf{D}.
\end{tcolorbox}

\end{minipage}

\caption{A sample from the \textit{Composition} aspect of \textit{Aesthetic Appreciation}.}
\label{fig:composition210}
\end{figure*}

\begin{figure*}[ht]
    \begin{tcolorbox}[colback=gray!5!white, colframe=gray!75!black,
    title=Prompt for \lconn{} Generation, boxrule=0.3mm, width=\textwidth, arc=3mm, auto outer arc=true]
    \scriptsize
\# ROLE

You are an expert AI Benchmark Analyst. Your task is to deconstruct a given question and its context to find the correct answer, select the best distractors, and state the core reasoning.
\newline
\newline
\# TASK

Analyze the provided JSON data, which contains ground-truth information, a \texttt{question}, and a list of potential \texttt{options}. Perform the following steps:
\begin{enumerate}
    \item Use the image to understand the ground truth. If an image is provided, the \texttt{explanation} text is supplementary.
    \item Read the \texttt{question} and evaluate all \texttt{options} against the image.
    \item Identify the single best \texttt{option} that is most strongly supported by the ground truth.
    \item From the remaining incorrect options, select the \textbf{three most plausible and confusing distractors}.
    \item Write a concise, one-sentence \texttt{reasoning} that explains \textit{why} the correct option is correct, based \textit{strictly} on the image.
\end{enumerate}
\ \newline
\# INPUT CONTEXT

The input is a single JSON object containing the raw data for a question. It contains the following structure:
\begin{itemize}
    \item \texttt{explanation}: A text string providing context for the scenario.
    \item \texttt{question}: The question to be analyzed.
    \item \texttt{options}: An array of strings, where one is the correct answer and the others are potential distractors.
\end{itemize}
\ \newline
\# OUTPUT FORMAT

Return a single JSON object with the following structure:
\begin{verbatim}
{
  "level": 3,
  "level_name": "Connotation",
  "question": "The original question text",
  "options": [
    {
      "option_text": "Text of the correct option",
      "is_correct": true
    },
    {
      "option_text": "Text of the first distractor",
      "is_correct": false
    },
    {
      "option_text": "Text of the second distractor",
      "is_correct": false
    },
    {
      "option_text": "Text of the third distractor",
      "is_correct": false
    }
  ],
  "reasoning": "The one-sentence explanation for why the correct answer is correct, 
                based on the provided ground truth."
}
\end{verbatim}
\ \newline
\# IMPORTANT RULES
\begin{itemize}
    \item The \texttt{question} in the output must be identical to the input \texttt{question}.
    \item The output \texttt{options} array must contain exactly \textbf{four} options: the single correct answer and the three most plausible distractors selected from the original list.
    \item Ensure exactly one option is marked as \texttt{is\_correct: true}.
    \item The \texttt{reasoning} is the most critical part. It must be clear and directly derivable from the image.
    \item Return ONLY the JSON object. Do not explain or add extra text.
\end{itemize}
\ \newline
Input:
\begin{verbatim}
{json_data}
\end{verbatim}
\end{tcolorbox}
    \caption{Prompt for \lconn{} generation on \bench{}.}
    \label{fig:prompt_l3analysis}
\end{figure*}

\begin{figure*}[ht]
    \begin{tcolorbox}[colback=gray!5!white, colframe=gray!75!black,
    title=Prompt for \lbridge{} Generation, boxrule=0.3mm, width=\textwidth, arc=3mm, auto outer arc=true]
    \scriptsize
\# ROLE

You are an expert AI Benchmark Crafter. You create questions that build on each other to test understanding.
\newline
\newline
\# TASK

You are given the analysis of a Level 3 question. Your task is to create a Level 2 (Comprehensive Understanding) question that logically precedes it.

The Level 2 question should address the "how" or "why" of the situation, bridging Level 1 facts and the Level 3 conclusion. It must be answerable using the image and help users understand the \texttt{reasoning} in the \texttt{level\_3\_analysis}.
\newline
\newline
\# INPUT CONTEXT

The input is a JSON object containing the following keys:
\begin{itemize}
    \item \texttt{explanation}: Supplementary text context. The primary ground truth is the image.
    \item \texttt{level\_3\_analysis}: The L3 QA and reasoning. Your L2 question must be a logical prerequisite for this.
\end{itemize}
\ \newline
\# OUTPUT FORMAT

Return a single JSON object for the Level 2 question:
\begin{verbatim}
{
  "level": 2,
  "level_name": "Semantic Bridge (Comprehensive Understanding)",
  "question": "Your newly generated Level 2 question",
  "options": [
    {
      "option_text": "The correct answer text",
      "is_correct": true
    },
    {
      "option_text": "A plausible but incorrect distractor",
      "is_correct": false
    },
    {
      "option_text": "Another plausible but incorrect distractor",
      "is_correct": false
    },
    {
      "option_text": "A third plausible but incorrect distractor",
      "is_correct": false
    }
  ]
}
\end{verbatim}

\# IMPORTANT RULES
\begin{itemize}
    \item The generated question and all options must be derived \textit{only} from the image.
    \item The question should focus on understanding the relationships, causes, or implications described in the provided context.
    \item \textbf{DISTRACTOR DIFFICULTY REQUIREMENTS FOR LEVEL 2 (CRITICAL):}
    \begin{itemize}
        \item Distractors must require \textbf{deliberate analysis} to rule out, not just superficial pattern matching.
        \item All distractors must be from the same \textbf{semantic/visual domain} as the correct answer and share overlapping features.
        \item Each distractor should be \textbf{partially correct}---it may align with part of the scene, concept, or logic---but contains a \textbf{crucial flaw}, such as:
        \begin{itemize}
            \item Oversimplification of a process or mechanism
            \item Misidentification of cause and effect
            \item Confusion between similar entities, directions, or states
            \item Overgeneralization from a local detail
        \end{itemize}
        \item Include at least one distractor that reflects a \textbf{common but incorrect heuristic}, such as:
        \begin{itemize}
            \item Selecting the most visually salient item regardless of function
            \item Assuming temporal sequence implies causality
            \item Mistaking correlation for explanation
        \end{itemize}
        \item Avoid options that are impossible, irrelevant, or rely on external knowledge.
        \item Test the ability to: differentiate superficial vs. deep correctness, resolve ambiguous cues, and recognize subtle misalignments.
    \end{itemize}
    \item Ensure there are exactly four options and only one is marked \texttt{is\_correct: true}.
    \item Return ONLY the JSON object. Do not explain or add extra text.
\end{itemize}
\{retry\_guidance\}
\newline
\newline
Input:
\begin{verbatim}
{
  "explanation": {explanation_text},
  "level_3_analysis": {level_3_data}
}
\end{verbatim}
\end{tcolorbox}
    \caption{Prompt for \lbridge{} generation on \bench{}.}
    \label{fig:prompt_l2generation}
\end{figure*}

\begin{figure*}[ht]
    \begin{tcolorbox}[colback=gray!5!white, colframe=gray!75!black,
    title=Prompt for \lperc{} Generation, boxrule=0.3mm, width=\textwidth, arc=3mm, auto outer arc=true]
    \scriptsize
\# ROLE

You are an expert AI Benchmark Crafter. You create questions that build on each other to test understanding.
\newline
\newline
\# TASK

You are given the analysis of a Level 3 question and the complete data for a generated Level 2 question. Your task is to create a \textbf{very simple} Level 1 (Perception) question that serves as the logical first step in this hierarchy.

The Level 1 question should ask about the \textbf{most prominent and easily verifiable element} in the image. The goal is to create a straightforward, entry-level question that almost any observer could answer easily. It should be significantly easier than the Level 2 question. It is the "what" that precedes the "how/why" of Level 2.
\newline
\newline
\# INPUT CONTEXT

The input is a JSON object containing the following keys:
\begin{itemize}
    \item \texttt{explanation}: A text string providing context for the scenario. If an image is part of the input, this explanation is supplementary information. The primary ground truth is defined by the image.
    \item \texttt{level\_3\_analysis}: A JSON object containing the Level 3 question, its correct answer, and the reasoning behind it. This provides the high-level context.
    \item \texttt{level\_2\_qa}: A JSON object containing the generated Level 2 question, its options, and the correct answer. Your Level 1 question should be a logical prerequisite for this question.
\end{itemize}
\ \newline
\# OUTPUT FORMAT

Return a single JSON object for the Level 1 question:
\begin{verbatim}
{
  "level": 1,
  "level_name": "Perception",
  "question": "Your newly generated, simple Level 1 question about 
               a prominent and central fact",
  "options": [
    {
      "option_text": "The correct factual answer",
      "is_correct": true
    },
    {
      "option_text": "A plausible but incorrect factual distractor",
      "is_correct": false
    },
    {
      "option_text": "Another plausible but incorrect factual distractor",
      "is_correct": false
    },
    {
      "option_text": "A third plausible but incorrect factual distractor",
      "is_correct": false
    }
  ]
}
\end{verbatim}

\# IMPORTANT RULES
\begin{itemize}
    \item The generated question and all options must be derived \textit{only} from the image.
    \item The question must be about a specific, observable, and \textbf{obvious} factual detail.
    \item The options must be plausible, with one clear correct answer based on the image.
    \item Ensure there are exactly four options, and only one is marked \texttt{is\_correct:~true}.
    \item Return ONLY the JSON object. Do not explain or add extra text.
\end{itemize}
\{retry\_guidance\}
\newline
\newline
Input:
\begin{verbatim}
{
  "explanation": {explanation_text},
  "level_3_analysis": {level_3_data},
  "level_2_qa": {level_2_data}
}
\end{verbatim}
\end{tcolorbox}
    \caption{Prompt for \lperc{} generation on \bench{}.}
    \label{fig:prompt_l1generation}
\end{figure*}

\begin{figure*}[ht]
    \begin{tcolorbox}[colback=gray!5!white, colframe=gray!75!black,
    title=Prompt for \lbridge{} $\rightarrow$ \lconn{} Validation, boxrule=0.3mm, width=\textwidth, arc=3mm, auto outer arc=true]
    \scriptsize
\# ROLE

You are a meticulous AI assistant specializing in logical and hierarchical analysis.
\newline
\newline
\# TASK

Evaluate if knowing the answer to "Level 2 QA" provides a foundational building block that helps in reasoning about or answering "Level 3 QA".
\newline
\newline
\textbf{CRITICAL REQUIREMENTS:}
\begin{enumerate}
    \item \textbf{Logical dependency}: Level 2 information must be DIRECTLY useful or necessary for Level 3
    \item \textbf{Difficulty progression}: Level 2 MUST be more objective, concrete, and simpler than Level 3
    \item \textbf{Hierarchical coherence}: Level 2 must provide intermediate knowledge that Level 3 builds upon
    \item \textbf{Complexity standard}: Level 2 should involve basic analysis/interpretation while Level 3 requires deeper reasoning, hidden meanings, or abstract concepts
\end{enumerate}

\textbf{VALIDATION STANDARDS:}
\begin{itemize}
    \item Level 2 questions should involve straightforward analysis or interpretation
    \item Level 3 questions should require complex reasoning, metaphorical understanding, or deeper insights
    \item There must be a clear logical connection where Level 2 knowledge helps answer Level 3
    \item If Level 2 is not noticeably simpler and more concrete than Level 3, validation should FAIL
\end{itemize}
\ \newline
\# REMEMBER

Your primary source of truth is the image.
\newline
\newline
\# QA Pairs

\textbf{Level 2 QA} (The foundational knowledge - should be more objective/simple)
\begin{itemize}
    \item Question: \{question\_l2\}
    \item Correct Answer: \{answer\_l2\}
\end{itemize}
\ \newline
\textbf{Level 3 QA} (The complex question that should build upon Level 2)
\begin{itemize}
    \item Question: \{question\_l3\}
    \item Correct Answer: \{answer\_l3\}
\end{itemize}
\ \newline
\# OUTPUT FORMAT

Respond with ONLY a JSON object with the following structure:
\begin{verbatim}
{
  "is_helpful": <boolean, true if Level 2 helps with Level 3 AND is significantly simpler, otherwise false>,
  "confidence": <float, your confidence in the "is_helpful" assessment from 0.0 to 1.0>,
  "reasoning": "<string, a brief explanation focusing on logical dependency and difficulty progression. 
                If false, explain why Level 2 is not sufficiently simpler or helpful>"
}
\end{verbatim}
\end{tcolorbox}
    \caption{Prompt for hierarchical validation on \bench{} (\lbridge{} $\rightarrow$ \lconn{}).}
    \label{fig:prompt_validation_l2_l3}
\end{figure*}

\begin{figure*}[ht]
    \begin{tcolorbox}[colback=gray!5!white, colframe=gray!75!black,
    title=Prompt for \lperc{} $\rightarrow$ \lbridge{} Validation, boxrule=0.3mm, width=\textwidth, arc=3mm, auto outer arc=true]
    \scriptsize
\# ROLE

You are a meticulous AI assistant specializing in logical and hierarchical analysis.
\newline
\newline
\# TASK

Evaluate if knowing the answer to "Level 1 QA" provides a foundational building block that helps in reasoning about or answering "Level 2 QA".
\newline
\newline
\textbf{CRITICAL REQUIREMENTS:}
\begin{enumerate}
    \item \textbf{Logical dependency}: Level 1 information must be DIRECTLY useful or necessary for Level 2
    \item \textbf{Difficulty progression}: Level 1 MUST be significantly more objective, concrete, and simpler than Level 2
    \item \textbf{Hierarchical coherence}: Level 1 must provide basic, factual knowledge that Level 2 builds upon
    \item \textbf{Objectivity standard}: Level 1 should focus on observable facts (colors, objects, numbers, basic actions) while Level 2 involves interpretation or analysis
\end{enumerate}
\ \newline
\textbf{VALIDATION STANDARDS:}
\begin{itemize}
    \item Level 1 questions should be answerable by direct observation
    \item Level 2 questions should require reasoning, interpretation, or analysis
    \item There must be a clear logical connection where Level 1 knowledge helps answer Level 2
    \item If Level 1 is not noticeably simpler and more objective than Level 2, validation should FAIL
\end{itemize}
\ \newline
\# REMEMBER

Your primary source of truth is the image.
\newline
\newline
\# QA Pairs

\textbf{Level 1 QA} (The foundational knowledge - should be most objective/simple)
\begin{itemize}
    \item Question: \{question\_l1\}
    \item Correct Answer: \{answer\_l1\}
\end{itemize}
\ \newline
\textbf{Level 2 QA} (The intermediate complexity question that should build upon Level 1)
\begin{itemize}
    \item Question: \{question\_l2\}
    \item Correct Answer: \{answer\_l2\}
\end{itemize}
\ \newline
\# OUTPUT FORMAT

Respond with ONLY a JSON object with the following structure:
\begin{verbatim}
{
  "is_helpful": <boolean, true if Level 1 helps with Level 2 AND is significantly simpler, otherwise false>,
  "confidence": <float, your confidence in the "is_helpful" assessment from 0.0 to 1.0>,
  "reasoning": "<string, a brief explanation focusing on logical dependency and difficulty progression.
                If false, explain why Level 1 is not sufficiently simpler or helpful>"
}
\end{verbatim}
\end{tcolorbox}
    \caption{Prompt for hierarchical validation on \bench{} (\lperc{} $\rightarrow$ \lbridge{}).}
    \label{fig:prompt_validation_l1_l2}
\end{figure*}

\clearpage
\begin{figure*}[ht]
    \begin{tcolorbox}[colback=gray!5!white, colframe=gray!75!black,
    title=Prompt for Level-1 Node Generation on the MCTS Tree, boxrule=0.3mm, width=\textwidth, arc=3mm, auto outer arc=true]
    \scriptsize
Your task is to generate a basic perception question based on the given context.
\newline
\newline
\textbf{Context:}
\begin{itemize}
    \item Target Level: \{target\_level\} - \{level\_description\}
    \item \{retry\_guidance\}
\end{itemize}
\ \newline
\textbf{Instructions:}
\begin{enumerate}
    \item Create a question that focuses on direct visual elements, objects, colors, positions, or basic attributes
    \item The question should be foundational and provide a building block for more complex reasoning
    \item Ensure the question is different from existing questions in the hierarchy
    \item The difficulty should be: \{difficulty\_guidance\}
    \item The answer MUST be concise ($\leq$ 30 words).
\end{enumerate}
\ \newline
You must respond with a JSON object in exactly this format:
\begin{verbatim}
{
  "question": "Your generated question here",
  "answer": "The correct answer",
  "reasoning": "Brief explanation of why this question fits level 1"
}
\end{verbatim}

Generate a level 1 basic perception question now.
\end{tcolorbox}
    \caption{Prompt for level-1 generation on data generation pipeline.}
    \label{fig:prompt_sft_l1_gen}
\end{figure*}

\begin{figure*}[ht]
    \begin{tcolorbox}[colback=gray!5!white, colframe=gray!75!black,
    title=Prompt for Level-2 Node Generation on the MCTS Tree, boxrule=0.3mm, width=\textwidth, arc=3mm, auto outer arc=true]
    \scriptsize
Your task is to generate a connection-level question based on the parent context.
\newline
\newline
\textbf{Parent Context:}
\begin{itemize}
    \item Parent Question: \{parent\_question\}
    \item Parent Answer: \{parent\_answer\}
    \item Target Level: \{target\_level\} - \{level\_description\}
    \item \{retry\_guidance\}
\end{itemize}
\ \newline
\textbf{Instructions:}
\begin{enumerate}
    \item Build upon the parent question/answer to create a more complex question
    \item Focus on relationships, connections, implications, or broader understanding
    \item The question should logically follow from the parent but require additional reasoning
    \item The difficulty should be: \{difficulty\_guidance\}
    \item The answer MUST be concise ($\leq$ 40 words).
    \item Ensure clear logical progression from the parent question
\end{enumerate}
\ \newline
You must respond with a JSON object in exactly this format:
\begin{verbatim}
{
  "question": "Your generated question here",
  "answer": "The correct answer",
  "reasoning": "Brief explanation of how this connects to the parent and why it fits level 2"
}
\end{verbatim}

Generate a level 2 connection question now.
\end{tcolorbox}
    \caption{Prompt for level-2 generation on data generation pipeline.}
    \label{fig:prompt_sft_l2_gen}
\end{figure*}

\begin{figure*}[ht]
    \begin{tcolorbox}[colback=gray!5!white, colframe=gray!75!black,
    title=Prompt for Level-3 Node Generation on the MCTS Tree, boxrule=0.3mm, width=\textwidth, arc=3mm, auto outer arc=true]
    \scriptsize
Your task is to generate a high-level reasoning question based on the parent context.
\newline
\newline
\textbf{Parent Context:}
\begin{itemize}
    \item Parent Question: \{parent\_question\}
    \item Parent Answer: \{parent\_answer\}
    \item Target Level: \{target\_level\} - \{level\_description\}
    \item \{retry\_guidance\}
\end{itemize}
\ \newline
\textbf{Instructions:}
\begin{enumerate}
    \item Build upon the parent question/answer to create a highly complex question
    \item Focus on abstract reasoning, inference, analysis, implications, or deep understanding
    \item The question should require sophisticated thinking beyond basic observation or connection
    \item The difficulty should be: \{difficulty\_guidance\}
    \item Ensure the question represents the pinnacle of reasoning complexity for this hierarchy
    \item The answer MUST be concise ($\leq$ 50 words).
\end{enumerate}
\ \newline
You must respond with a JSON object in exactly this format:
\begin{verbatim}
{
  "question": "Your generated question here",
  "answer": "The correct answer",
  "reasoning": "Brief explanation of how this builds on the parent and why it requires high-level reasoning"
}
\end{verbatim}

Generate a level 3 high-level reasoning question now.
\end{tcolorbox}
    \caption{Prompt for level-3 generation on data generation pipeline.}
    \label{fig:prompt_sft_l3_gen}
\end{figure*}

\begin{figure*}[ht]
    \begin{tcolorbox}[colback=gray!5!white, colframe=gray!75!black,
    title=Prompt for Evaluation of New Nodes on the MCTS Tree, boxrule=0.3mm, width=\textwidth, arc=3mm, auto outer arc=true]
    \scriptsize
You are an expert evaluator for hierarchical visual question-answer datasets. Evaluate the quality of the target Q\&A pair.
\newline
\newline
\textbf{Context (for reference only):}
\begin{itemize}
    \item Previous Level \{parent\_level\} Question: \{parent\_question\}
    \item Previous Level \{parent\_level\} Answer: \{parent\_answer\}
\end{itemize}
\ \newline
\textbf{Target Q\&A to Evaluate:}
\begin{itemize}
    \item Level \{child\_level\} Question: \{child\_question\}
    \item Level \{child\_level\} Answer: \{child\_answer\}
    \item Expected Level: \{child\_level\} (\{level\_description\})
\end{itemize}
\ \newline
\textbf{Core Evaluation Criteria (3 key dimensions):}
\begin{enumerate}
    \item \textbf{Logical Coherence}: Is the question internally consistent and does the answer logically follow from the question?
    \item \textbf{Difficulty Appropriateness}: Does the question match the cognitive demands of Level \{child\_level\}? Does it demonstrate the expected depth of thinking?
    \item \textbf{Image Alignment}: Do both the question and answer accurately reflect and align with the provided image content?
\end{enumerate}
\ \newline
\textbf{Evaluation Rules:}
\begin{itemize}
    \item Evaluate each dimension on a 0--1.0 scale with precision
    \item Consider overall quality holistically
    \item Focus reasoning on identifying current limitations and areas needing improvement
    \item High scores (0.8+) for excellent Q\&A pairs, medium scores (0.5--0.7) for adequate ones, low scores (<0.5) for problematic ones
\end{itemize}
\ \newline
You must respond with JSON in exactly this format:
\begin{verbatim}
{
  "quality_score": 0.0-1.0,
  "reasoning": "Identify specific issues and areas that need improvement (focus on limitations, not strengths)"
}
\end{verbatim}

Provide your evaluation now.
\end{tcolorbox}
    \caption{Prompt for evaluator on data generation pipeline.}
    \label{fig:prompt_sft_eval_hierarchical}
\end{figure*}

%% file: image-and-table/statistics.tex
\begin{table}[H]
\centering
\caption{Statistics of \bench{} across three task families and fifteen fine-grained aspects.}
\label{tab:statistics}
\resizebox{\textwidth}{!}{ 
\begin{tabular}{l ccccc cccc cccccc c}
\toprule

& \multicolumn{5}{c}{\textbf{Implication Understanding (400)}} 
& \multicolumn{4}{c}{\textbf{Aesthetic Appreciation (350)}} 
& \multicolumn{6}{c}{\textbf{Affective Reasoning (300)}} 
& \multirow{2}{*}{\textbf{Total}} \\ 

\cmidrule(lr){2-6} \cmidrule(lr){7-10} \cmidrule(lr){11-16}

& 
\textbf{Metaphor} & \textbf{Symbolism} & \textbf{Contrast} & \textbf{Exaggeration} & \textbf{Dislocation} & 
\textbf{Color} & \textbf{Composition} & \textbf{Font} & \textbf{Graphics} & 
\textbf{Joy} & \textbf{Affection} & \textbf{Wonder} & \textbf{Anger} & \textbf{Fear} & \textbf{Sadness} & 
\\ 

\cmidrule(lr){1-17}

\textbf{\#Samples} & 
319 & 21 & 22 & 22 & 16 & 
37 & 122 & 97 & 94 &        
25 & 83 & 47 & 31 & 81 & 33 & 
1050
\\ 

\bottomrule
\end{tabular}
}
\end{table}

%% file: image-and-table/lora_results.tex
\begin{table}[t]
\centering
\caption{Performance of \lora{} on \bench{}. $+\Delta$ denotes gain over ``base'' setting.}
\label{tab:lora_results}
\resizebox{\columnwidth}{!}{%
\begin{tabular}{cccc cccc cccc c}
    \toprule
    \multicolumn{4}{c}{\textbf{Implication Understanding}} & 
    \multicolumn{4}{c}{\textbf{Aesthetic Appreciation}} & 
    \multicolumn{4}{c}{\textbf{Affective Reasoning}} & 
    \multirow{2}{*}{\textbf{Score}} \\
    \cmidrule(lr){1-4} \cmidrule(lr){5-8} \cmidrule(lr){9-12}
    \bm{\aperc{}} & \bm{\abridge{}} & \bm{\aconn{}} & \bm{\afull{}} & 
    \bm{\aperc{}} & \bm{\abridge{}} & \bm{\aconn{}} & \bm{\afull{}} & 
    \bm{\aperc{}} & \bm{\abridge{}} & \bm{\aconn{}} & \bm{\afull{}} & \\
    \midrule
    
    \rowcolor{gray!15}
    \multicolumn{13}{c}{\textbf{``Base'' Setting}} \\
    \midrule
    90.50 & 86.75 & 61.25 & 50.00 & 90.29 & 74.29 & 66.57 & 46.57 & 93.33 & 83.67 & 63.00 & 45.67 & 47.41 \\
    \midrule
    
    \rowcolor{gray!15}
    \multicolumn{13}{c}{\textbf{``Context'' Setting}} \\
    \midrule
    90.50 & 87.25 & 71.75 & 58.25{\scriptsize\colorbox{gray!15}{$+8.25$}} & 90.29 & 76.57 & 85.14 & 58.57{\scriptsize\colorbox{gray!15}{$+12.00$}} & 93.33 & 87.00 & 73.33 & 59.67{\scriptsize\colorbox{gray!15}{$+14.00$}} & 58.83{\scriptsize\colorbox{gray!15}{$+11.42$}} \\
    \bottomrule
\end{tabular}
}
\end{table}

%% file: image-and-table/general_benchmarks.tex
\begin{table}[t]
\centering
\caption{Detailed Results on general benchmarks. The best score in each metric is \textbf{in-bold}.}
\label{tab:general_benchmarks}
\resizebox{\textwidth}{!}{%
\begin{tabular}{l ccccc ccc ccccccc c}
    \toprule
    \multirow{2}{*}{\textbf{Model}} & \multicolumn{5}{c}{\textbf{MMBench}} & \multicolumn{3}{c}{\textbf{HallusionBench}} & \multicolumn{7}{c}{\textbf{MMStar}} & \textbf{MMMU} \\
    \cmidrule(lr){2-6} \cmidrule(lr){7-9} \cmidrule(lr){10-16} \cmidrule(lr){17-17}
     & \textbf{CN(CC)} & \textbf{CN(Dev)} & \textbf{EN(Dev)} & \textbf{RU(Dev)} & \textbf{Avg.} & \textbf{qAcc} & \textbf{fAcc} & \textbf{aAcc} & \textbf{CP} & \textbf{FP} & \textbf{IR} & \textbf{LR} & \textbf{ST} & \textbf{MA} & \textbf{Avg.} & \textbf{Acc} \\
    \midrule
    Qwen3-VL-4B-Instruct & \textbf{68.82} & 81.36 & \textbf{83.93} & 77.42 & 77.88 & \textbf{36.48} & 36.71 & \textbf{60.25} & 77.03 & 60.22 & 69.43 & 53.26 & 43.72 & 41.02 & 57.45 & 48.56 \\
    \lora{} & 68.43 & \textbf{82.13} & 83.76 & \textbf{78.18} & \textbf{78.18} & 36.04 & \textbf{36.99} & 59.62 & \textbf{78.06} & \textbf{60.39} & \textbf{71.60} & \textbf{62.02} & \textbf{50.39} & \textbf{65.80} & \textbf{64.71} & \textbf{51.78} \\
    \bottomrule
\end{tabular}
}
\end{table}

%% file: image-and-table/detailed_results.tex
\begin{table*}[t]
\centering
\caption{Overall results on \bench{} with ``context'' setting.}
\label{tab:detailed_results}
\resizebox{\textwidth}{!}{%
\begin{tabular}{l r cccc cccc cccc c}
    \toprule
    \multirow{2}{*}{\textbf{Model}} & \multirow{2}{*}{\textbf{Model Size}} & \multicolumn{4}{c}{\textbf{Implication Understanding}} & \multicolumn{4}{c}{\textbf{Aesthetic Appreciation}} & \multicolumn{4}{c}{\textbf{Affective Reasoning}} & \multirow{2}{*}{\textit{\textbf{Score}}} \\
    \cmidrule(lr){3-6} \cmidrule(lr){7-10} \cmidrule(lr){11-14}
     & & \bm{\aperc{}} & \bm{\abridge{}} & \bm{\aconn{}} & \bm{\afull{}} & \bm{\aperc{}} & \bm{\abridge{}} & \bm{\aconn{}} & \bm{\afull{}} & \bm{\aperc{}} & \bm{\abridge{}} & \bm{\aconn{}} & \bm{\afull{}} & \\
    \midrule
    \multicolumn{15}{l}{\textbf{Basic Reference}} \\
    \midrule
    GPT-4o & - & 95.50 & 89.75 & 76.50 & 65.00 & 95.43 & 82.29 & 87.71 & 72.86 & 91.33 & 86.00 & 80.67 & 66.67 & 68.18 \\
    \midrule
    \multicolumn{15}{l}{\textbf{Open-Source MLLMs}} \\
    \midrule
    Qwen3-VL-Instruct & 4B & 86.75 & 85.50 & 70.75 & 54.50 & 90.57 & 72.86 & 74.00 & 53.14 & 90.33 & 86.00 & 74.33 & 57.67 & 55.10 \\
    Qwen3-VL-Instruct & 8B & 93.50 & 90.00 & 74.75 & 62.75 & 91.71 & 74.00 & 82.57 & 59.43 & 94.33 & 89.00 & 76.00 & 64.67 & 62.28 \\
    LLaVA-1.6 & 7B & 81.75 & 68.00 & 54.50 & 32.75 & 79.14 & 43.71 & 50.00 & 18.29 & 92.00 & 65.00 & 30.00 & 18.67 & 23.24 \\
    LLaVA-1.6 & 13B & 84.75 & 80.25 & 63.00 & 44.75 & 84.86 & 57.71 & 52.00 & 29.14 & 94.33 & 78.00 & 37.67 & 27.33 & 33.74 \\
    Deepseek-VL2-tiny & MoE 1B/3B & 88.25 & 65.25 & 55.75 & 34.25 & 89.71 & 47.71 & 60.57 & 27.43 & 93.33 & 68.33 & 33.00 & 23.00 & 28.23 \\
    Deepseek-VL2 & MoE 4.5B/27B & 93.75 & 84.25 & 67.50 & 53.75 & 95.14 & 59.43 & 54.00 & 33.43 & 96.33 & 83.67 & 62.00 & 52.00 & 46.39 \\
    Gemma3 & 4B & 76.50 & 78.25 & 63.50 & 40.75 & 68.86 & 65.14 & 82.57 & 35.43 & 87.00 & 75.00 & 50.00 & 33.00 & 36.39 \\
    Gemma3 & 12B & 87.50 & 88.00 & 74.50 & 57.00 & 82.86 & 72.57 & 84.86 & 50.00 & 90.67 & 86.00 & 74.33 & 59.67 & 55.56 \\
    InternVL3.5 & 4B & 82.50 & 80.75 & 66.75 & 46.00 & 82.86 & 65.43 & 64.00 & 36.86 & 91.00 & 82.33 & 79.00 & 60.33 & 47.73 \\
    InternVL3.5 & 8B & 82.00 & 84.75 & 66.00 & 47.25 & 84.00 & 70.57 & 71.14 & 45.43 & 86.00 & 81.67 & 68.00 & 50.33 & 47.67 \\
    Phi-4-Multimodal-Instruct & 6B & 90.25 & 84.75 & 68.00 & 54.50 & 90.29 & 61.71 & 50.86 & 32.86 & 90.00 & 86.33 & 59.67 & 45.00 & 44.12 \\
    Phi-3.5-Vision-Instruct & 4B & 84.25 & 84.00 & 71.50 & 51.25 & 88.29 & 63.43 & 64.00 & 40.00 & 91.33 & 81.33 & 64.67 & 49.33 & 46.86 \\
    \bottomrule
\end{tabular}
}
\end{table*}

%% file: image-and-table/other_model.tex
\begin{table}[t]
\centering
\caption{Performance on Larger and Heterogeneous Models. $+\Delta$ denotes gain over the model before instruction tuning.}
\label{tab:other_model}
\resizebox{\textwidth}{!}{%
\begin{tabular}{l cccc cccc cccc c}
    \toprule
    \multirow{2}{*}{\textbf{Model}} & \multicolumn{4}{c}{\textbf{Implication Understanding}} & \multicolumn{4}{c}{\textbf{Aesthetic Appreciation}} & \multicolumn{4}{c}{\textbf{Affective Reasoning}} & \multirow{2}{*}{\textit{\textbf{Score}}} \\
    \cmidrule(lr){2-5} \cmidrule(lr){6-9} \cmidrule(lr){10-13}
     & \bm{\aperc{}} & \bm{\abridge{}} & \bm{\aconn{}} & \bm{\afull{}} & \bm{\aperc{}} & \bm{\abridge{}} & \bm{\aconn{}} & \bm{\afull{}} & \bm{\aperc{}} & \bm{\abridge{}} & \bm{\aconn{}} & \bm{\afull{}} & \\
    \midrule
    \multicolumn{14}{l}{\textbf{Larger Model}} \\
    \midrule
    Qwen3-VL-8B-Instruct & 93.50 & 89.50 & 59.50 & 50.75 & 91.71 & 73.43 & 63.43 & 44.00 & 94.33 & 84.67 & 60.00 & 48.00 & 47.58 \\
    Qwen3-VL-8B-Bridge & 93.75 & 91.25 & 62.25 & 55.00{\scriptsize\colorbox{gray!15}{$+4.25$}} & 91.71 & 78.29 & 70.00 & 52.57{\scriptsize\colorbox{gray!15}{$+8.57$}} & 94.33 & 87.33 & 60.33 & 49.00{\scriptsize\colorbox{gray!15}{$+1.00$}} & 52.19{\scriptsize\colorbox{gray!15}{$+4.61$}} \\
    \midrule
    \multicolumn{14}{l}{\textbf{Heterogeneous Model}} \\
    \midrule
    LLaVA-1.6-7B & 81.75 & 58.00 & 40.25 & 18.75 & 79.14 & 36.86 & 33.14 & 9.43 & 92.00 & 58.00 & 19.33 & 12.00 & 13.39 \\
    LLaVA-1.6-7B-Bridge & 82.00 & 75.25 & 52.50 & 33.25{\scriptsize\colorbox{gray!15}{$+14.50$}} & 75.71 & 43.43 & 47.14 & 15.71{\scriptsize\colorbox{gray!15}{$+6.28$}} & 92.33 & 61.33 & 22.00 & 14.00{\scriptsize\colorbox{gray!15}{$+2.00$}} & 20.99{\scriptsize\colorbox{gray!15}{$+7.60$}} \\
    \bottomrule
\end{tabular}
}
\end{table}

%% file: image-and-table/data_generation_comparision.tex
\begin{table}[t]
\centering
\caption{Comparison between Qwen3-VL-4B-Direct and Qwen3-VL-4B-Bridge on \bench{}.}
\label{tab:other_data}
\resizebox{\textwidth}{!}{%
\begin{tabular}{l cccc cccc cccc c}
    \toprule
    \multirow{2}{*}{\textbf{Model}} & \multicolumn{4}{c}{\textbf{Implication Understanding}} & \multicolumn{4}{c}{\textbf{Aesthetic Appreciation}} & \multicolumn{4}{c}{\textbf{Affective Reasoning}} & \multirow{2}{*}{\textit{\textbf{Score}}} \\
    \cmidrule(lr){2-5} \cmidrule(lr){6-9} \cmidrule(lr){10-13}
     & \bm{\aperc{}} & \bm{\abridge{}} & \bm{\aconn{}} & \bm{\afull{}} & \bm{\aperc{}} & \bm{\abridge{}} & \bm{\aconn{}} & \bm{\afull{}} & \bm{\aperc{}} & \bm{\abridge{}} & \bm{\aconn{}} & \bm{\afull{}} & \\
    \midrule
    Qwen3-VL-4B-Direct & 89.50 & 87.00 & 60.25 & 48.75 & 88.57 & 74.57 & 64.57 & 44.00 & 92.33 & 83.67 & 58.67 & 43.33 & 45.36 \\
    Qwen3-VL-4B-Bridge & 90.50 & 86.75 & 61.25 & 50.00 & 90.29 & 74.29 & 66.57 & 46.57 & 93.33 & 83.67 & 63.00 & 45.67 & 47.41 \\
    \addlinespace[1pt]
    \hdashline[2pt/2pt]
    \addlinespace[1pt]
    \rowcolor{gray!15} $\Delta$ & \textbf{+1.00} & -0.25 & \textbf{+1.00} & \textbf{+1.25} & \textbf{+1.72} & -0.28 & \textbf{+2.00} & \textbf{+2.57} & \textbf{+1.00} & \textbf{+0.00} & \textbf{+4.33} & \textbf{+2.34} & \textbf{+2.05} \\
    \bottomrule
\end{tabular}
}
\end{table}